\documentclass[12pt]{article}

\usepackage{scicite}

\usepackage{times}

\usepackage{amsmath,amssymb}
\usepackage{graphicx}
\usepackage{algpseudocode}
\usepackage{algorithm}

\usepackage{bm}
\usepackage{tikz}
\DeclareMathOperator{\E}{\mathbb{E}}

\usepackage{comment}

\usepackage{subcaption}
\usepackage{pstool}
\usepackage[percent]{overpic}

\newenvironment{sciabstract}{%
\begin{quote} \bf}
{\end{quote}}

\newcounter{lastnote}

\title{Entropic maximization form intelligence of biological and robots, Swarm policy for robots driven by entropic maximization policy in ants swarms} 

\title{Micro-scale exploration driven by invisible macro-scale state to achieve evolutionary goal in ants and robots}

\title{Ant three has multiple phase simultaneously}

\title{Multi-phases swarm intelligence in biological swarm and robots}
\title{Unified diversity-driven-force of entropy maximization in physics, biology and intelligence}
\title{Unifying Collective Intelligence in Biological Swarms, Physical Systems, and Robotic Swarms by Entropy Maximization}
\title{Unifying Collective Intelligence in Ant Swarms and Robotic Swarms by Entropy Maximization principle in Physical system}
\title{Unified Collective Intelligence in Stochastic Behavior of Ant Swarms and Robotic Swarms by Entropy Maximization Principle in Physical System}
\title{Unified Stochastic mechanism of Ant, Physical and Robotic Swarms}
\title{A Unified Stochastic Mechanism Underlying Collective Behavior in Ants, Physical Systems, and Robotic Swarms}

\author
{ Lianhao Yin $^{1\ast}$, Haiping Yu $^{2}$, Pascal Spino $^{1}$ Daniela Rus $^{1}$\\
\\
\normalsize{$^{1}$ Computer Science and Artificial Intelligence Laboratory, MIT, MA, US}\\
\normalsize{$^{2}$ The International Institute for Industrial Environmental Economics, Lund University, Sweden}\\
\\
\normalsize{$^\ast$E-mail: lianhao, spino, rus@csail.mit.edu, haiping.h.yu@gmail.com}
}

\date{}

\usepackage{xr}
\begin{document}

\baselineskip24pt

\maketitle

\begin{sciabstract}
Biological swarms, such as ant colonies, achieve collective goals through decentralized and stochastic individual behaviors. Similarly, physical systems composed of gases, liquids, and solids exhibit random particle motion governed by entropy maximization, yet do not achieve collective objectives. Despite this analogy, no unified framework exists to explain the stochastic behavior in both biological and physical systems. Here, we present empirical evidence from \textit{Formica polyctena} ants that reveals a shared statistical mechanism underlying both systems: maximization under different energy function constraints. We further demonstrate that robotic swarms governed by this principle can exhibit scalable, decentralized cooperation, mimicking physical phase-like behaviors with minimal individual computation. These findings established a unified stochastic model linking biological, physical, and robotic swarms, offering a scalable principle for designing robust and intelligent swarm robotics.

\end{sciabstract}

\section{Introduction} \label{sec: introduction}
Biological swarms, such as ant colonies, exhibit seemingly random individual actions during foraging, object transport, and colony migration, while the collective swarm achieves intelligent behavior. In contrast, individual physical particles in gases, liquids, and solids also move randomly, but collectively they do not pursue a common goal driven by intrinsic motivation—such as transporting objects or moving in a preferential direction. Instead, matter particles exhibit collective physical properties such as phase, temperature, and pressure.
Noise provides several important benefits to collective systems. It can facilitate alignment in locust swarms \cite{Yates2009-we}, induce transitions between disordered and ordered phases in mobile insect groups \cite{Escudero2010-mb}, and enhance global human coordination in network experiments \cite{shirado2017locally}.
To date, no unified principle exists to explain the stochastic behavior observed in both systems. In general, in biological swarms, various simplified physical models have been proposed to explain collective behaviors, for example, the Ising model for cooperative transport \cite{Feinerman2018-ke}, the Vicsek model to describe the flocking of birds \cite{Toner1995-kk}, the dynamic model for bacteria turbulence \cite{Dunkel2013-mc} and human panic escape \cite{Helbing2000-qw}, and the quantification of collective state \cite{Tunstrom2013-yt}. 
Algorithmic approaches have also been used to model flocking and swarming, incorporating gradient, consensus, and group objective terms~\cite{Olfati-Saber2006-xb}, as well as neural network models such as the ring attractor network to explain order formation in locust swarms~\cite{Sayin2025-rh}. Specifically, several modeling methods have been employed to represent noise and investigate its role in swarms. A simple approach is to add white noise to the action \cite{vicsek1995novel}. Another approach introduces stochastic behavior by random sampling of peer perception \cite{bode2010making}. In addition, Brownian motion has been incorporated into differential equations to model alignment dynamics \cite{degond2015phase}.
However, these models often impose external coordination or lack a unified explanation for stochasticity across both physical and biological domains. Inspired by D'Arcy Thompson’s view that scientific understanding often arises from identifying analogies rather than causalities~\cite{Thompson1945-kk}, our goal is not to reduce biological behavior to physical mechanics, but rather to uncover a shared principle underlying both.

Therefore, one of the motivations of the paper is to understand the distribution randomness of biological swarms in constrained environments through tracking the motion of ants in a constrained environment, and to explore the randomness in multi-robot systems in which each robot is considered as a particle to find a relation to randomness in physical particles.  

Current multi-robot systems require communication to each other and a central controller to achieve collective goal or behavior, for example, formation control \cite{Rubenstein2014-vn}, self-reproduction \cite{zykov2005self}, and moving an object using particle robots \cite{Li2019-mp}. When the number of robots increases, the speed and ability of the system to achieve the collective goal, e.g. moving an object, decrease \cite{Li2019-mp}. This is due to that the demands for accurate coordination and consensus increase \cite{Yim2007-so, Spinos2017-fg} in compromise of robustness. The programming of centralized, multi-robot systems for various tasks is also difficult due to the complexity of algorithmic and mechanical coordination between units \cite{Liu2023-nj,Spinos2017-fg}. However, experimental robotic swarms such as manorobots \cite{Miskin2020-hh}, ant-like robots \cite{Rubenstein2014-vn}, magnetic-driven microrobots \cite{Gardi2022-fy}, and granular matter controlled by the magnetic field \cite{Kaiser2017-jz} instead utilize a fully decentralized control approach with minimum individual computation to allow scaling to larger numbers of robots with higher collective capabilities. %
Therefore, the second motivation of this paper is to replicate the stochastic action of ant swarms, enabling a fully decentralized and scalable robot swarm to achieve a common goal while necessitating minimum computation by each individual robot.   
\begin{figure}[h!]
    \centering
    \begin{subfigure}[b]{\textwidth}
    \includegraphics[width=\textwidth]{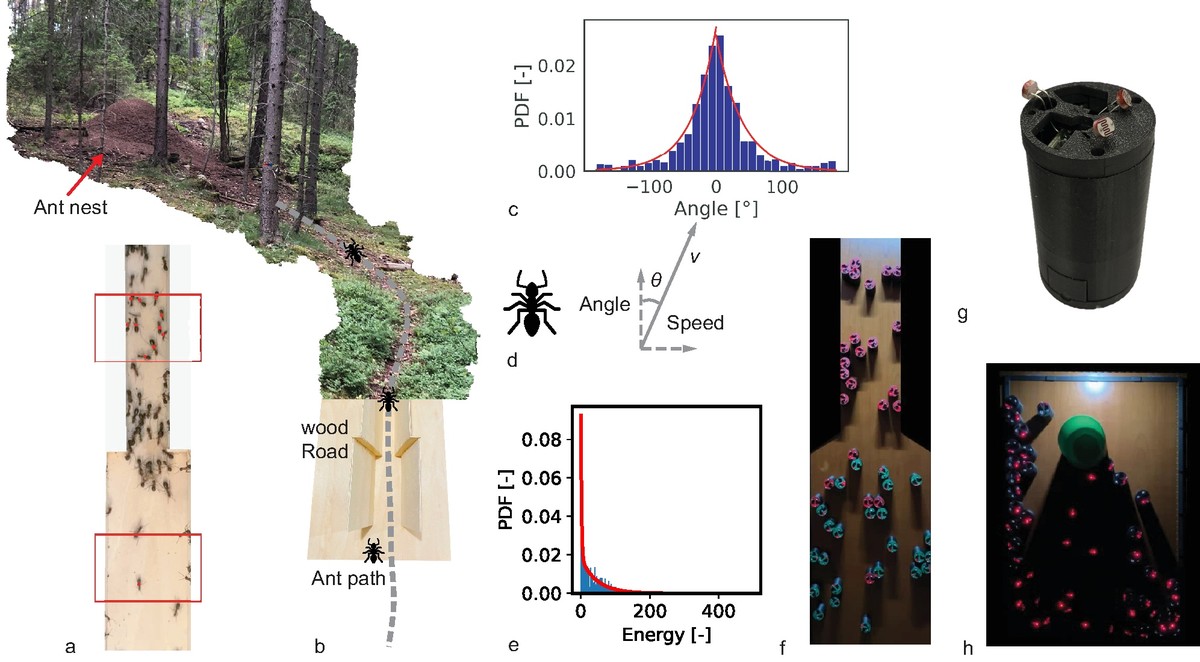}
    \end{subfigure}
    \caption{(a) We track the ants on the up and low side of the wood road, which is marked by red squares. (b) We put the wood road on the ant path between the ant nest and the forage areas. (c) The probability density of the angle rate of the ant. The red is fitted with an exponential distribution. (d) Illustrate how the speed and angle are defined. (e) The probability density function of energy of an ant, which is defined as ($E=\frac{1}{2}mv^2$), where $m$,$v$ are the mass and speed of the ant, respectively. (f) The setting of robotic experiments for narrow road crossing. (g) Illustration of the robot hardware. (h) The setting of robotic experiments for moving object.}
    \label{fig: field exp settings overview}
\end{figure}
\section{Ant swarm experiments} \label{sec: biological exp}
To investigate stochastic behavior in the ant swarm, we measured the speed and steering angle of biological ants (\textit{Formica polyctena}, one of the Formica species) in a constrained environment.
\textit{Formica polyctena} has been shown to sense both pheromone and colored light to navigate to the nest~\cite{kiepenheuer1968farbunterscheidungsvermogen}. A wooden road was constructed (Fig.~\ref{fig:road dim}) and placed within the natural flow path of the ant that leads to the nest (Fig.~\ref{fig: field exp settings overview}) serving as the main study location. This area has the highest density of ants in traffic. A top-down camera was used to record the ant flow (Fig.~\ref{fig: field exp settings overview}). The wooden road was placed within the ant nest to ensure a homogeneous pheromone coating on its surface prior to experiments. The length of the wooden road was chosen to maintain uninterrupted ant flow. The width of the road matched the original flow path to preserve natural traffic conditions and prevent ants from entering other paths. Previous experiment with different ant species show that the bidirectional ratio of the ant flow does not significantly influence the flow rate and the crowdedness of ants in congestion~\cite{Poissonnier2019-uf}. Accordingly, our experiments did not treat the bidirectional ratio as an experimental variable.

\begin{figure}[htbp] %
\centering
\begin{minipage}[b]{0.25\textwidth}
\begin{subfigure}{\textwidth}
\includegraphics[width=\textwidth]{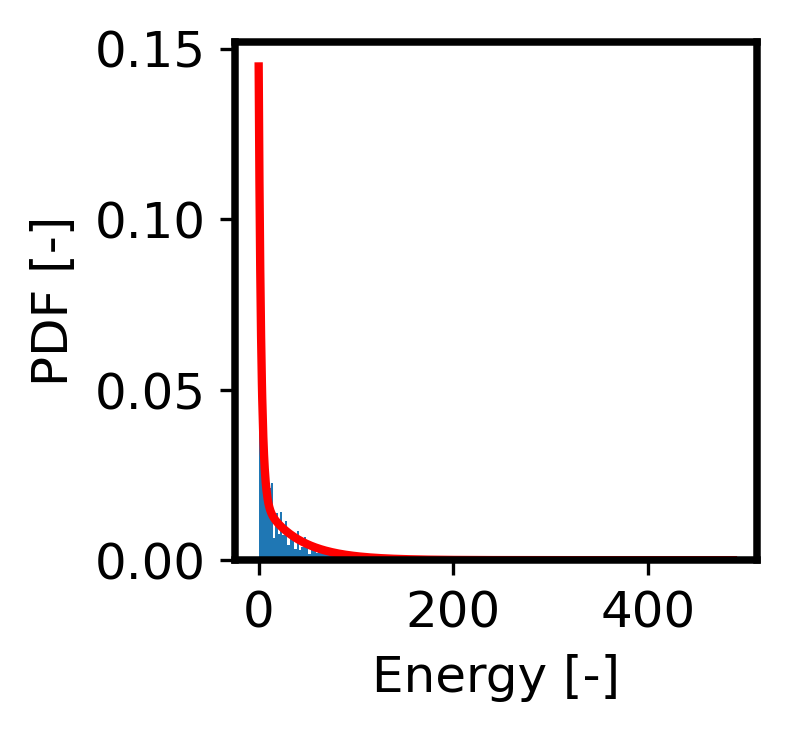}
\caption{\textnormal{ks: 0.052, p: 0.99}}
\label{fig: bio: T different speed}
\end{subfigure}
\end{minipage}
\begin{minipage}[b]{0.25\textwidth}
\begin{subfigure}{\textwidth}
\includegraphics[width=\textwidth]{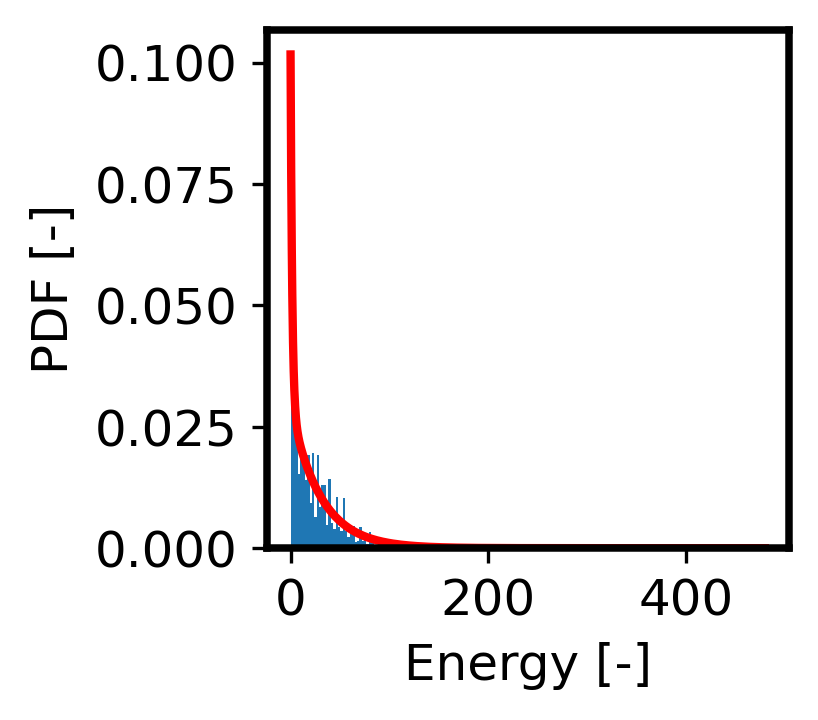}
\caption{\textnormal{ks: 0.030, p: 1.00}}
\label{fig: bio: T different speed}
\end{subfigure}
\end{minipage}
\begin{minipage}[b]{0.25\textwidth}
\begin{subfigure}{\textwidth}
\includegraphics[width=\textwidth]{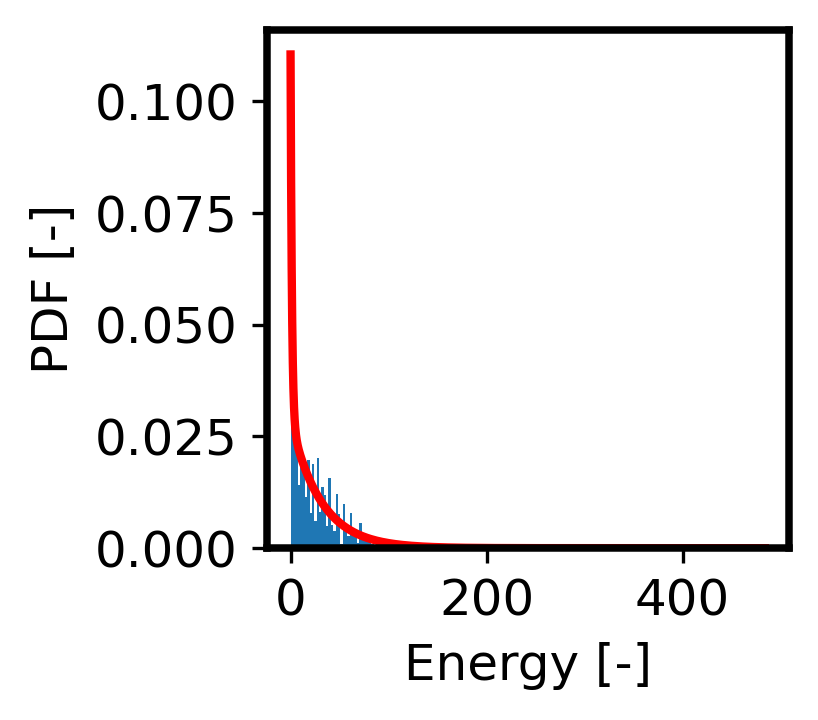}
\caption{\textnormal{ks: 0.028, p: 1.00}}
\label{fig: bio: T different speed}
\end{subfigure}
\end{minipage}
\begin{minipage}[b]{0.25\textwidth}
\begin{subfigure}{\textwidth}
\includegraphics[width=\textwidth]{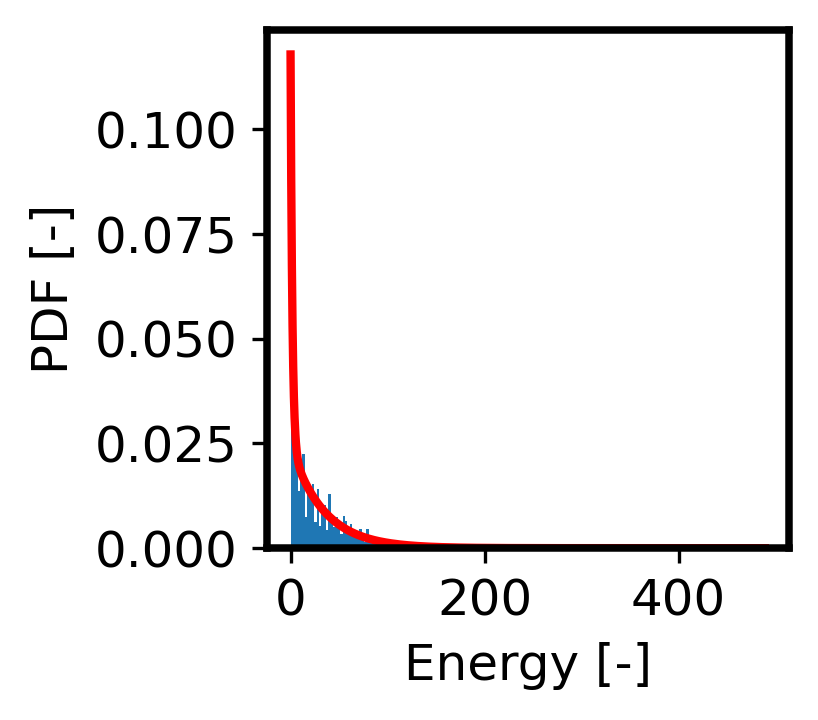}
\caption{\textnormal{ks: 0.029, p: 1.00}}
\label{fig: bio: expoential distribution of steering angle at 93 energy}
\end{subfigure}
\end{minipage}
\begin{minipage}[b]{0.25\textwidth}
\begin{subfigure}{\textwidth}
\includegraphics[width=\textwidth]{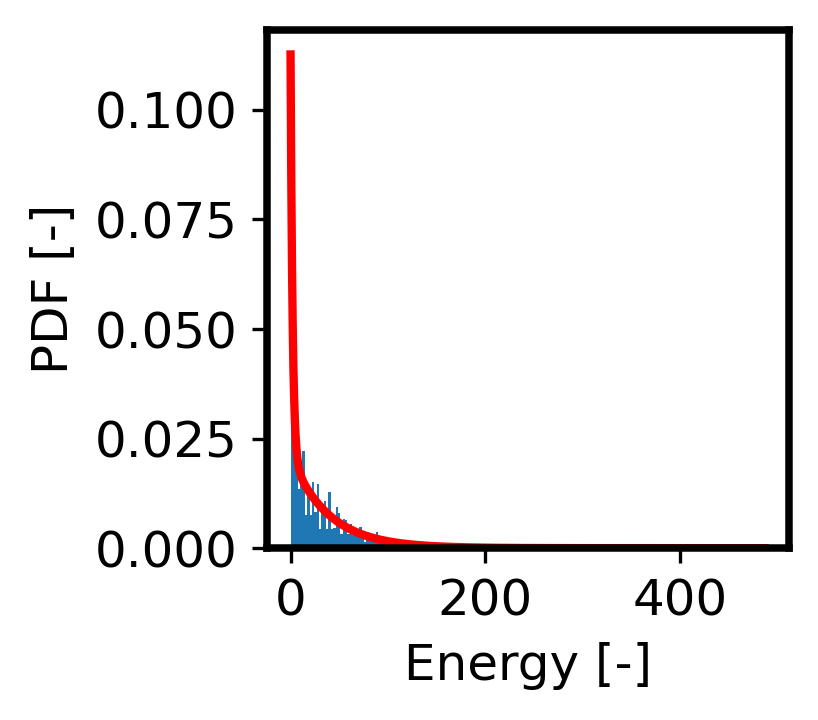}
\caption{\textnormal{ks: 0.034, p: 1.00}}
\label{fig: bio: expoential distribution of steering angle at 146 energy}
\end{subfigure}
\end{minipage}
\begin{minipage}[b]{0.25\textwidth}
\begin{subfigure}{\textwidth}
\includegraphics[width=\textwidth]{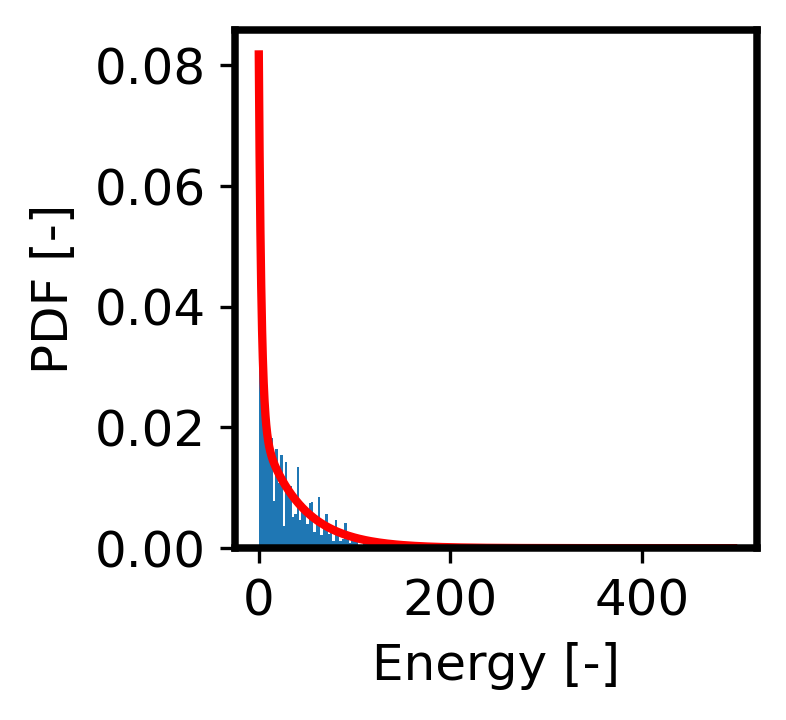}
\caption{\textnormal{ks: 0.028, p: 0.99}}
\label{fig: bio: expoential distribution of steering angle at 146 energy}
\end{subfigure}
\end{minipage}
\begin{minipage}[b]{0.25\textwidth}
\begin{subfigure}{\textwidth}
\includegraphics[width=\textwidth]{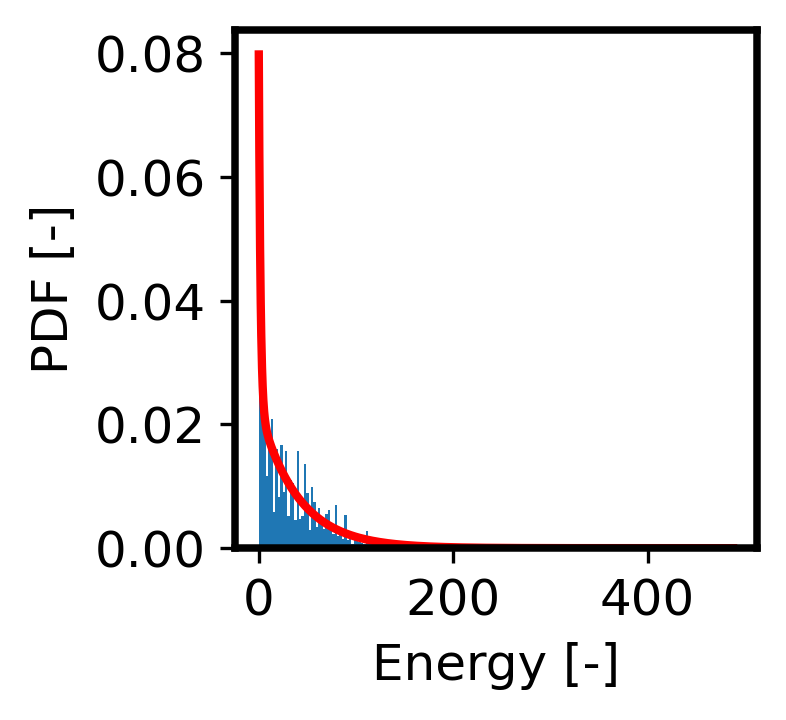}
\caption{\textnormal{ks: 0.035, p: 0.99}}
\label{fig: bio: expoential distribution of steering angle at 146 energy}
\end{subfigure}
\end{minipage}
\begin{minipage}[b]{0.25\textwidth}
\begin{subfigure}{\textwidth}
\includegraphics[width=\textwidth]{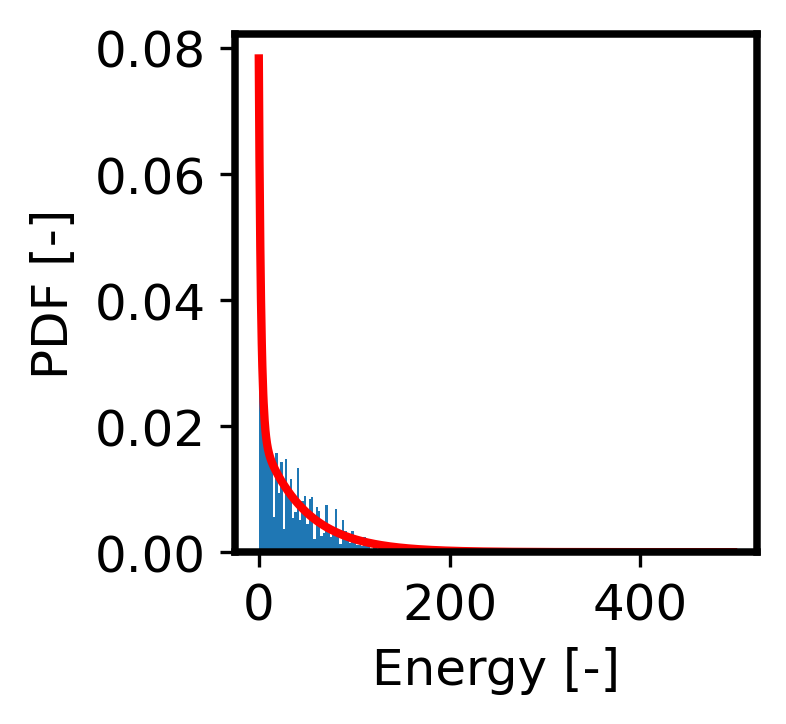}
\caption{\textnormal{ks: 0.024, p: 0.99}}
\label{fig: bio: expoential distribution of steering angle at 146 energy}
\end{subfigure}
\end{minipage}
\begin{minipage}[b]{0.25\textwidth}
\begin{subfigure}{\textwidth}
\includegraphics[width=0.25\textwidth]{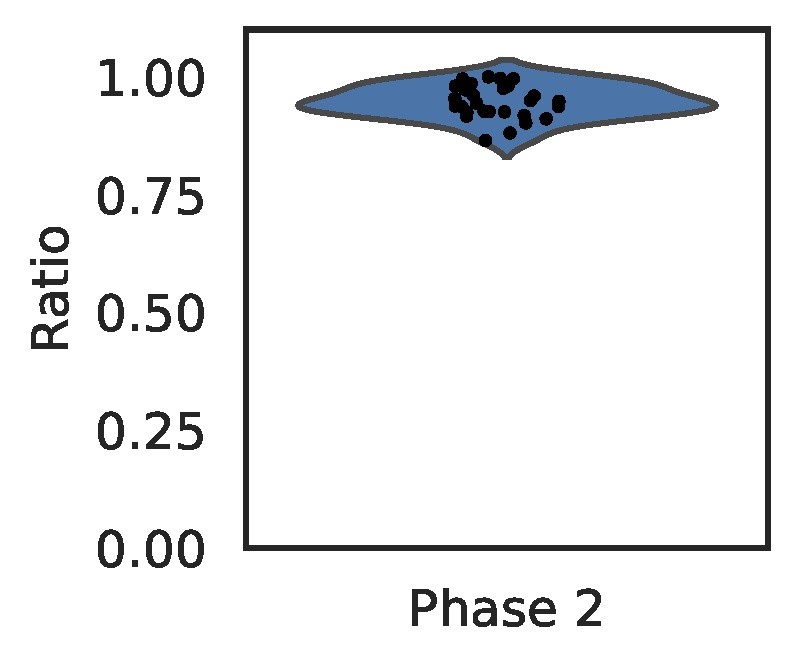}%
\caption{\textnormal{popmean=0.94. p: 1.00}}
\label{fig: bio: expoential distribution of steering angle at 146 energy}
\end{subfigure}
\end{minipage}
\vfill
\begin{minipage}[b]{0.25\textwidth}
\begin{subfigure}{\textwidth}
\includegraphics[width=0.25\textwidth]{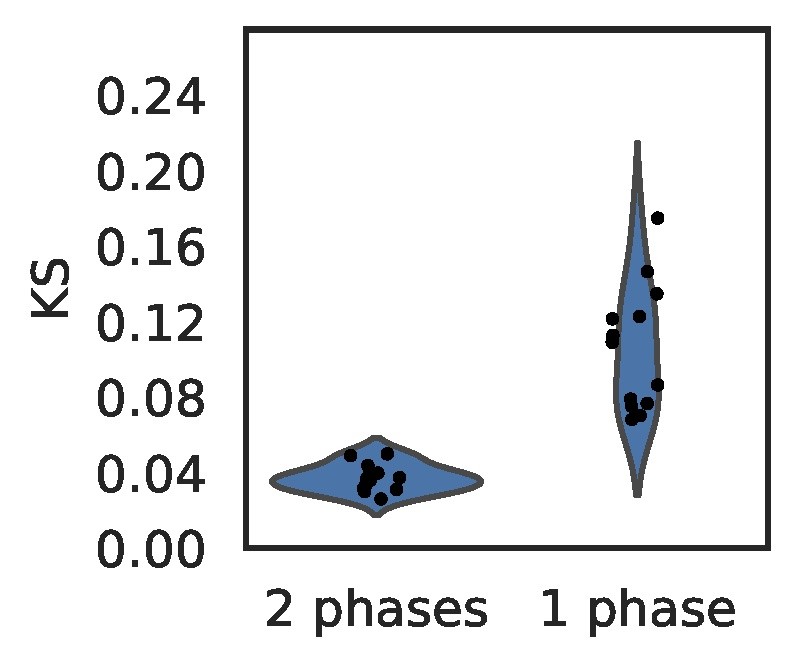}
\caption{\textnormal{p-value=$7e^{-8}$ in t-test}}
\end{subfigure} 
\end{minipage}
\begin{minipage}[b]{0.4\textwidth} %
\begin{subfigure}{\textwidth}
\includegraphics[width=\textwidth]{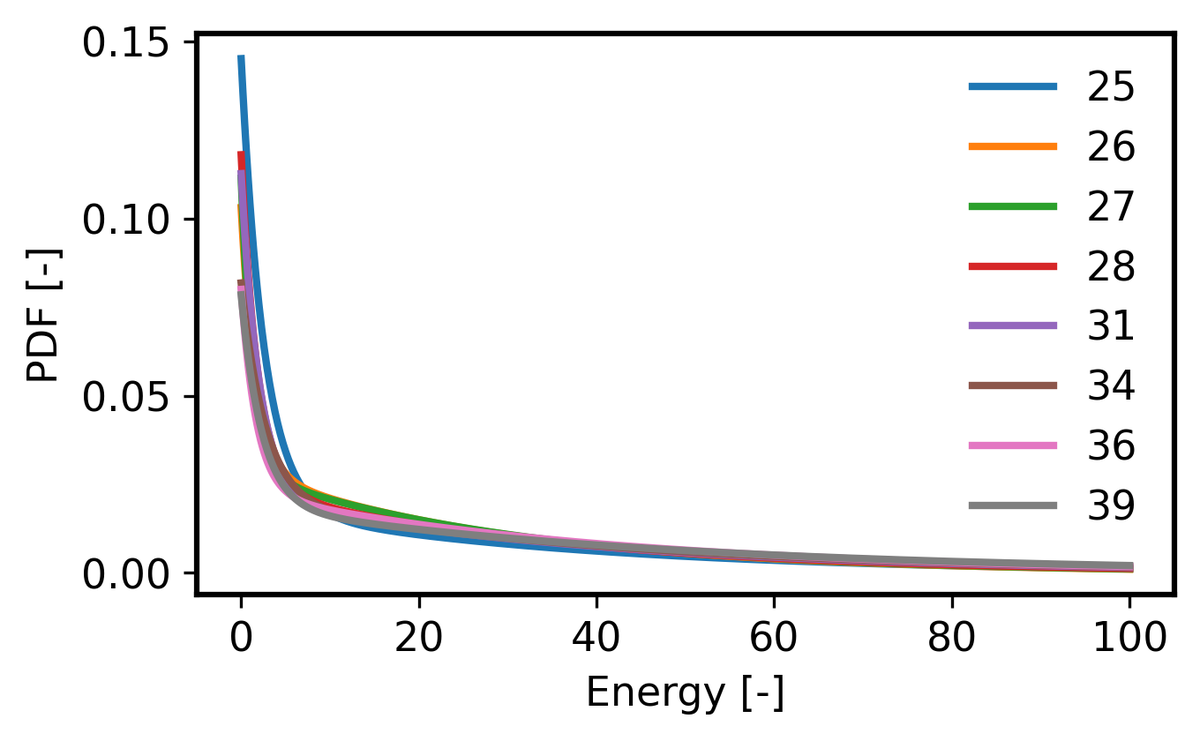}
\textbf{(k)}
\end{subfigure}
\end{minipage}
\caption{The speed distribution ($p(v^2_i) = \phi \frac{1}{kT_1}e^{-\frac{1}{kT_1}v^{2}_{i}} + (1-\phi) \frac{1}{kT_2} e^{-\frac{1}{kT_2}v^{2}_{i}}$ in red). (a-h) Energy level at  25, 26, 27, 28, 31, 34, 36, 39 respectively, (g) Probability density curve at different energy level. Unit:$1/10^10$ J, (j) KS of 1 phase versus 2 phases fitting, (i) Ratio of ant at upper and lower side in phase 2}
\label{fig: bio: speed distribution different speed}
\end{figure}

\section{The stochastic behaviour in ant swarm} \label{sec: principle of entropy maximization}
The analysis of individual ant trajectories, based on the tracking method described in Section~\ref{sec: method: video analysis and tracking}, shows that both the energy distribution ($E = \frac{v^2}{2}$, where $v$ is the speed) and the distribution of the steering angle rate ($\theta$) follow exponential forms consistent with energy-based statistical models.
Specifically, the probability density function of energy takes the form $p(v^2_i) = \phi \frac{1}{kT_1}e^{-\frac{1}{kT_1}v^{2}{i}} + (1-\phi) \frac{1}{kT_2} e^{-\frac{1}{kT_2}v^{2}{i}}$, and that of the steering angle rate takes the form $p(\theta) = \frac{1}{kT_{\theta}(v)} e^{-\frac{1}{kT_{\theta}(v)}\theta}$ (Fig.~\ref{fig: field exp settings overview}c,e).

\begin{figure}[htp]
\begin{minipage}[b]{0.3\textwidth}
\begin{subfigure}{\textwidth}
\includegraphics[width=\textwidth]{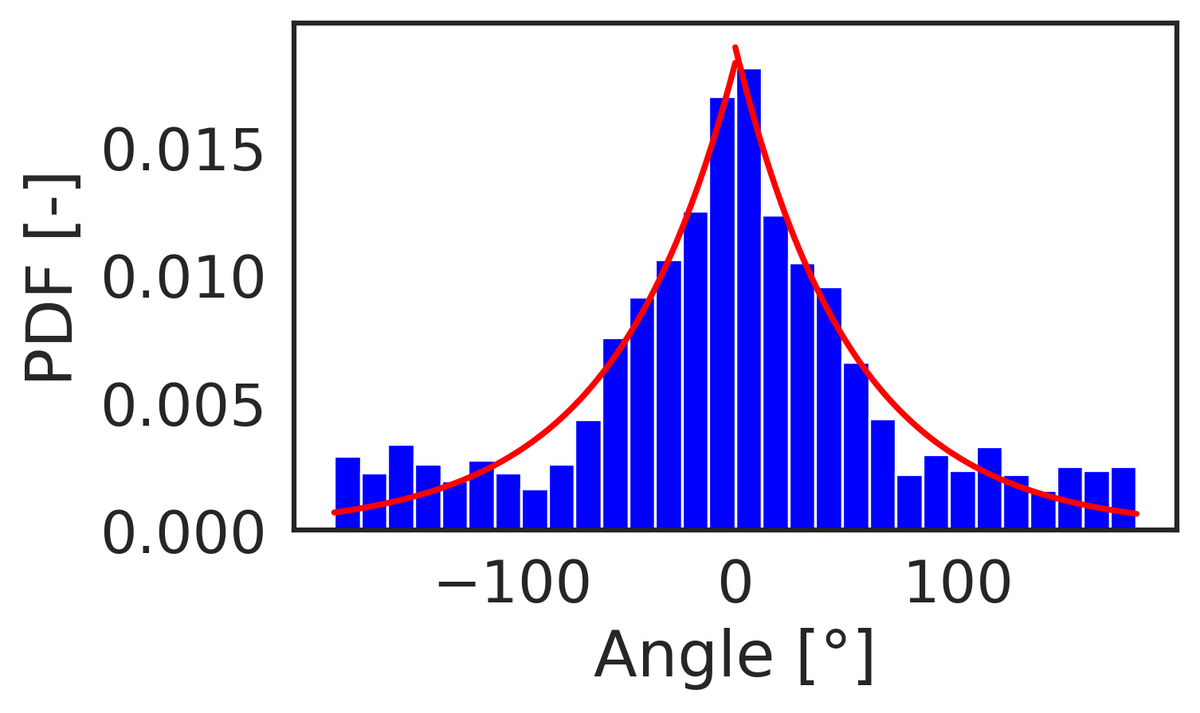}
\caption{}
\label{fig: bio: T different speed}
\end{subfigure}
\end{minipage}
\begin{minipage}[b]{0.3\textwidth}
\begin{subfigure}{\textwidth}
\includegraphics[width=\textwidth]{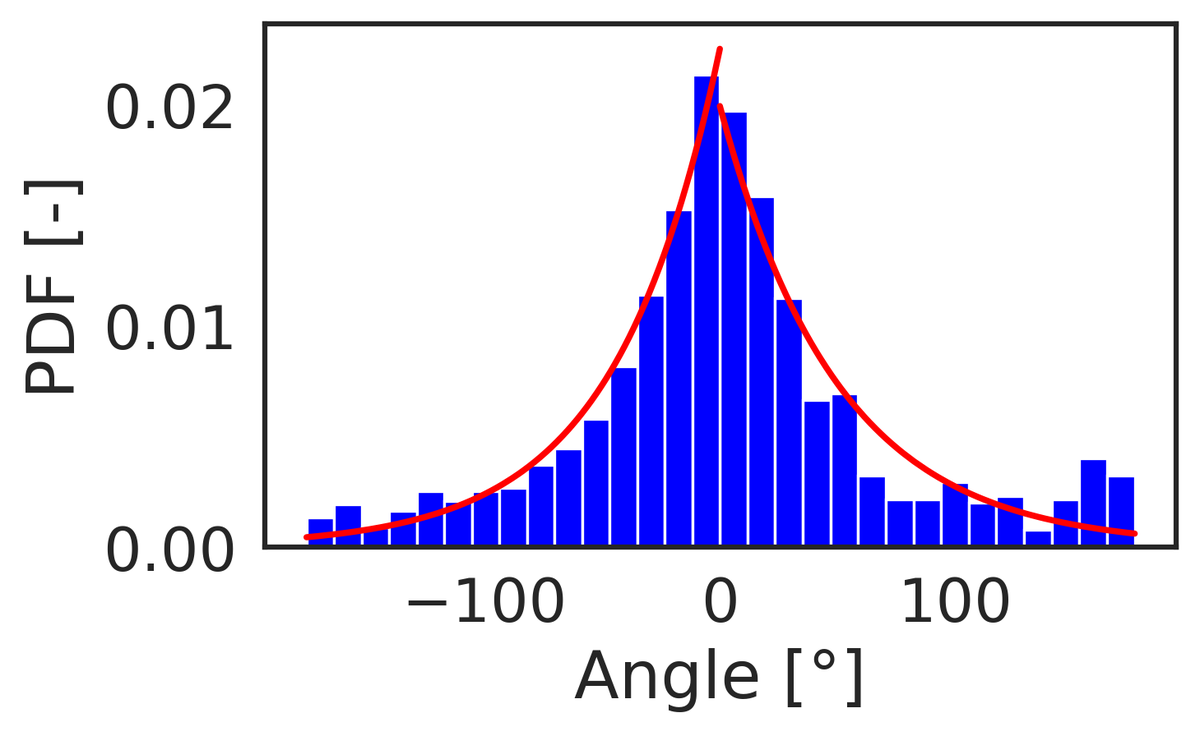}
\caption{}
\label{fig: bio: T different speed}
\end{subfigure}
\end{minipage}
\begin{minipage}[b]{0.3\textwidth}
\begin{subfigure}{\textwidth}
\includegraphics[width=\textwidth]{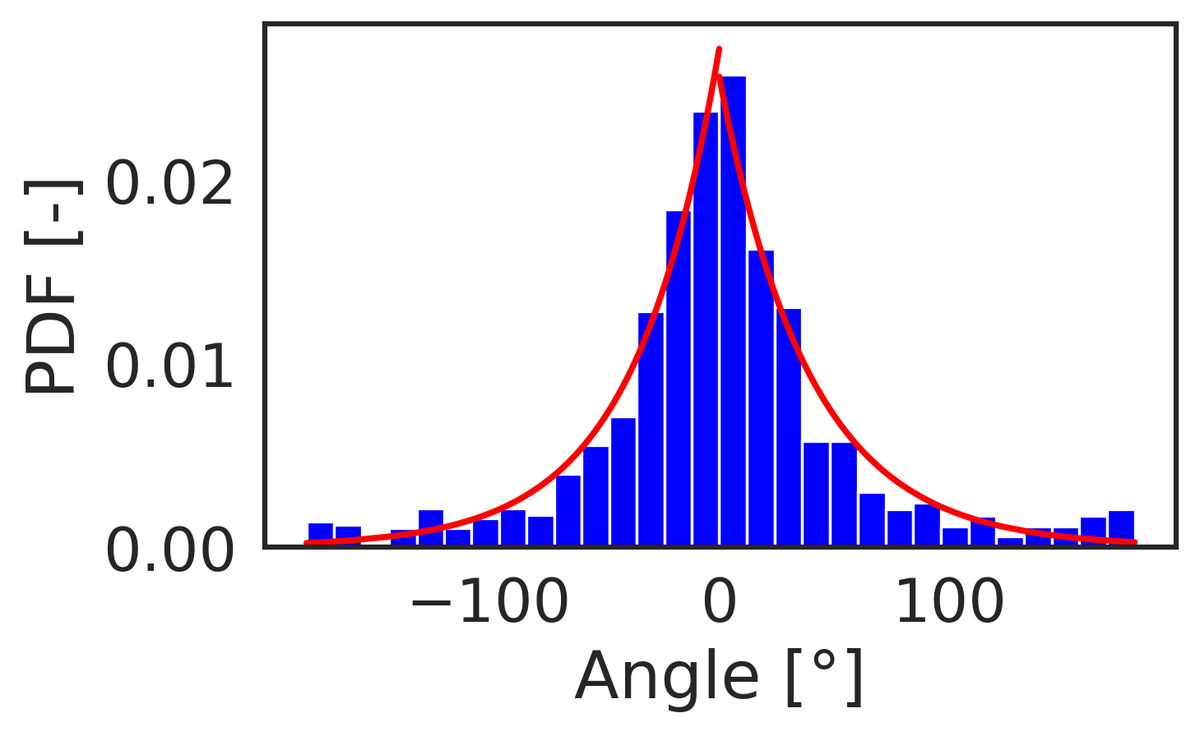}
\caption{}
\label{fig: bio: expoential distribution of steering angle at 93 energy}
\end{subfigure}
\end{minipage}
\begin{minipage}[b]{0.3\textwidth}
\begin{subfigure}{\textwidth}
\includegraphics[width=\textwidth]{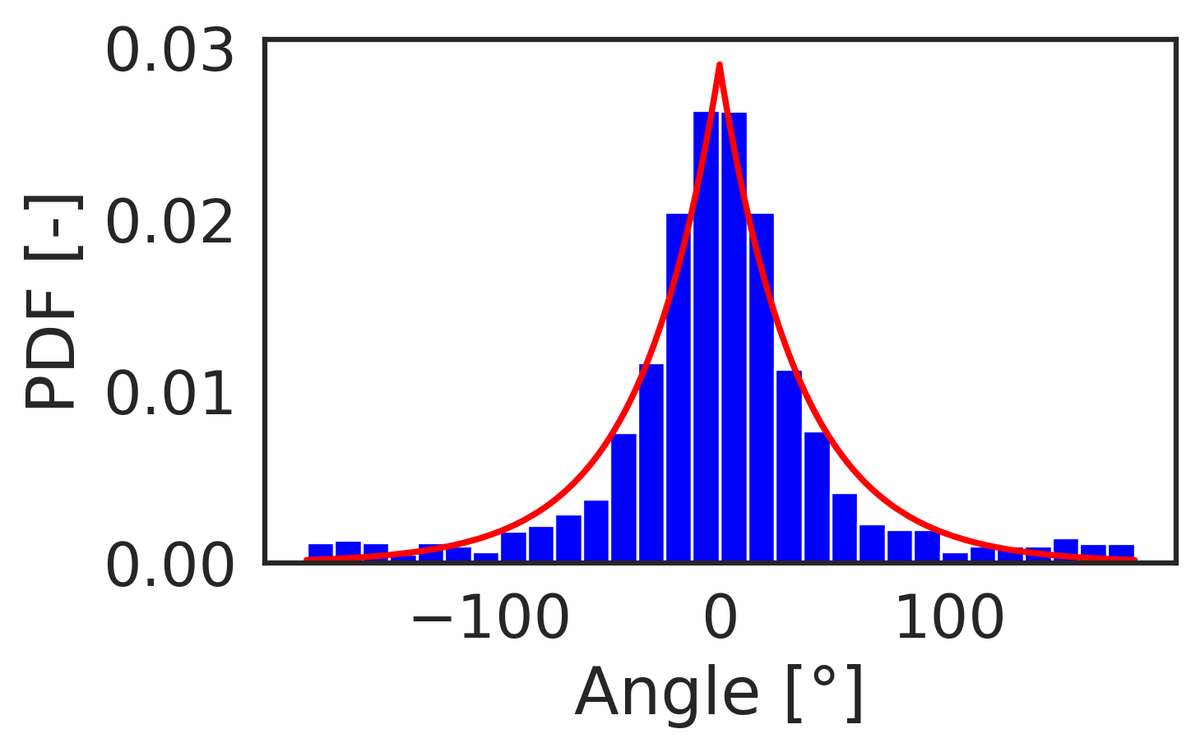}
\caption{}
\label{fig: bio: expoential distribution of steering angle at 146 energy}
\end{subfigure}
\end{minipage}
\begin{minipage}[b]{0.3\textwidth}
\begin{subfigure}{\textwidth}
\includegraphics[width=\textwidth]{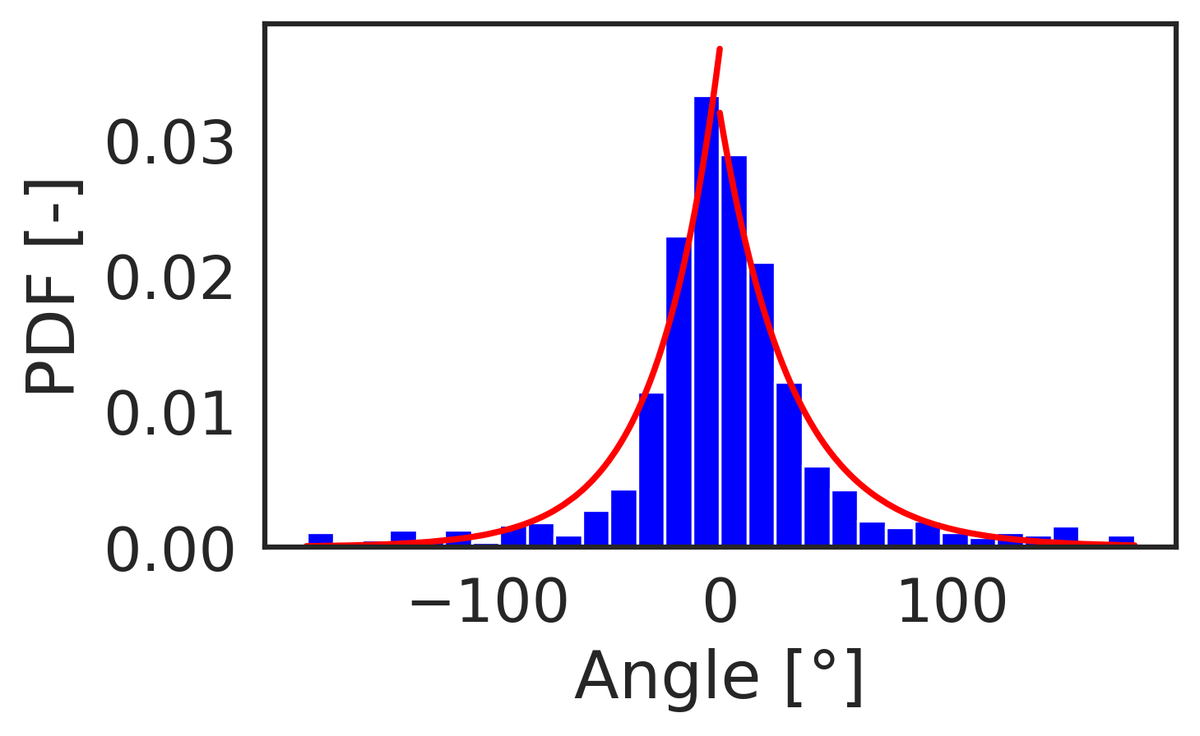}
\caption{}
\label{fig: bio: expoential distribution of steering angle at 146 energy}
\end{subfigure}
\end{minipage}
\begin{minipage}[b]{0.3\textwidth}
\begin{subfigure}{\textwidth}
\includegraphics[width=\textwidth]{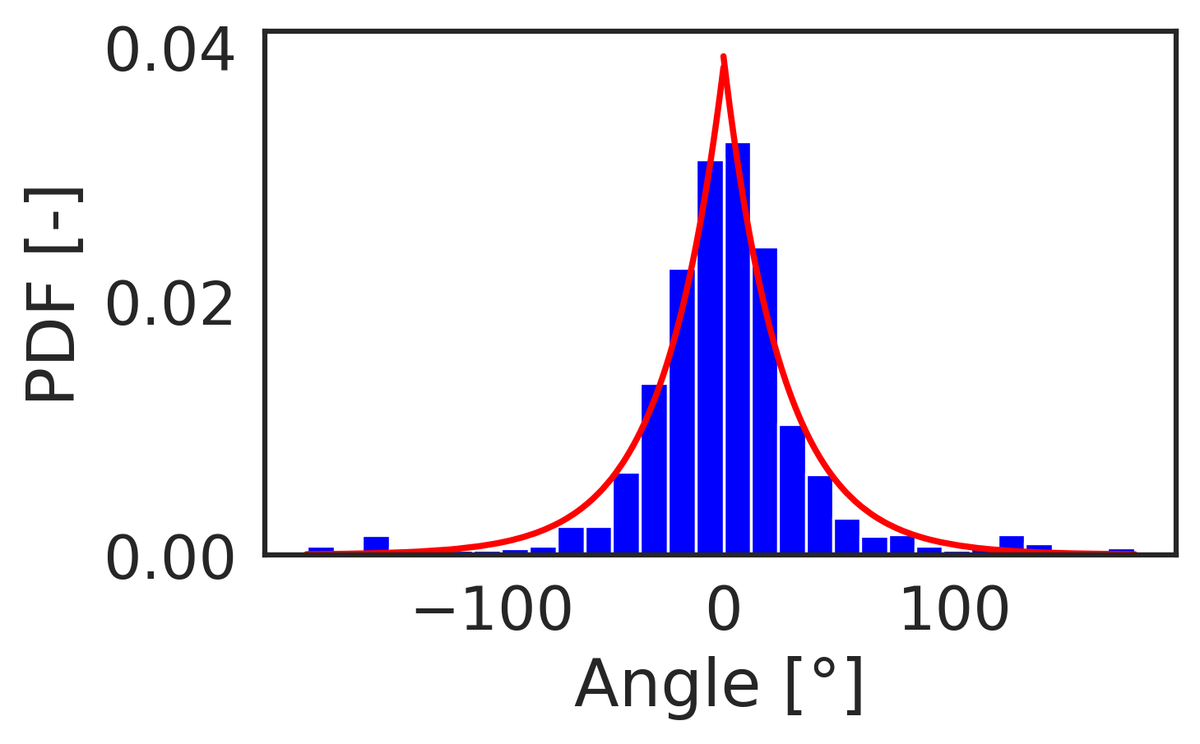}
\caption{}
\label{fig: bio: expoential distribution of steering angle at 146 energy}
\end{subfigure}
\end{minipage}
\begin{minipage}[b]{0.3\textwidth}
\begin{subfigure}{\textwidth}
\includegraphics[width=\textwidth]{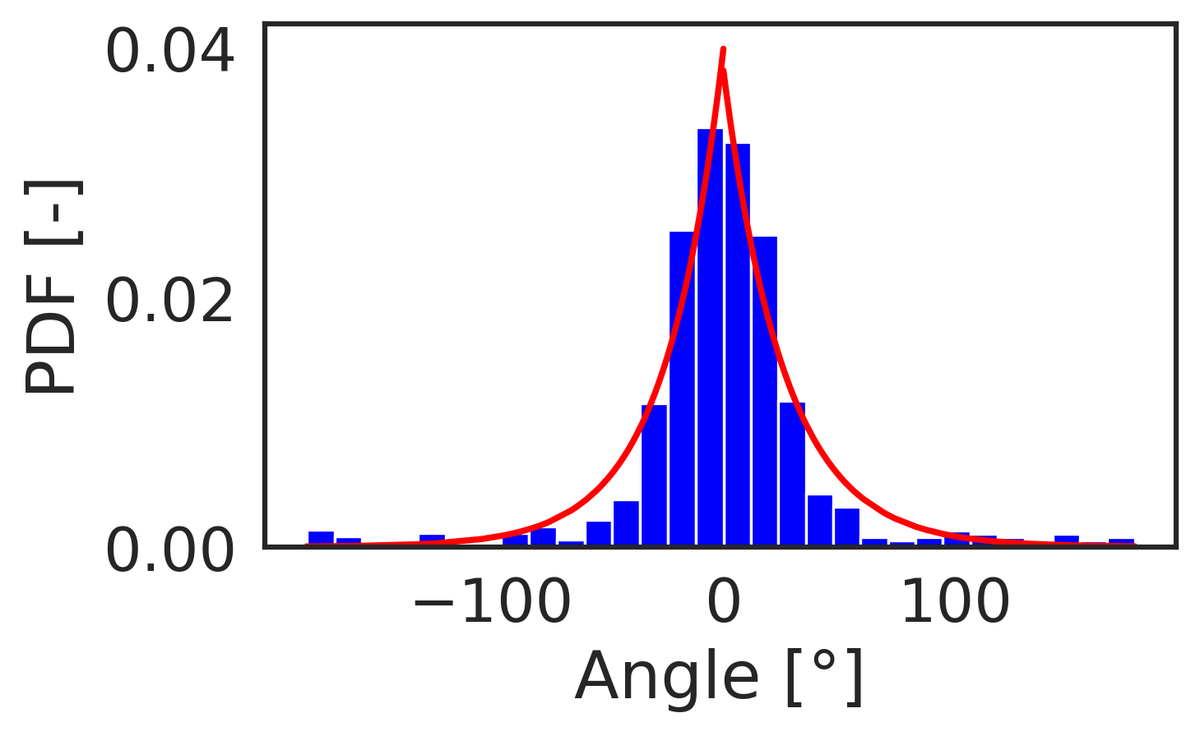}
\caption{}
\label{fig: bio: expoential distribution of steering angle at 146 energy}
\end{subfigure}
\end{minipage}
\hfill
\begin{minipage}[b]{0.3\textwidth}
\begin{subfigure}{\textwidth}
\includegraphics[width=\textwidth]{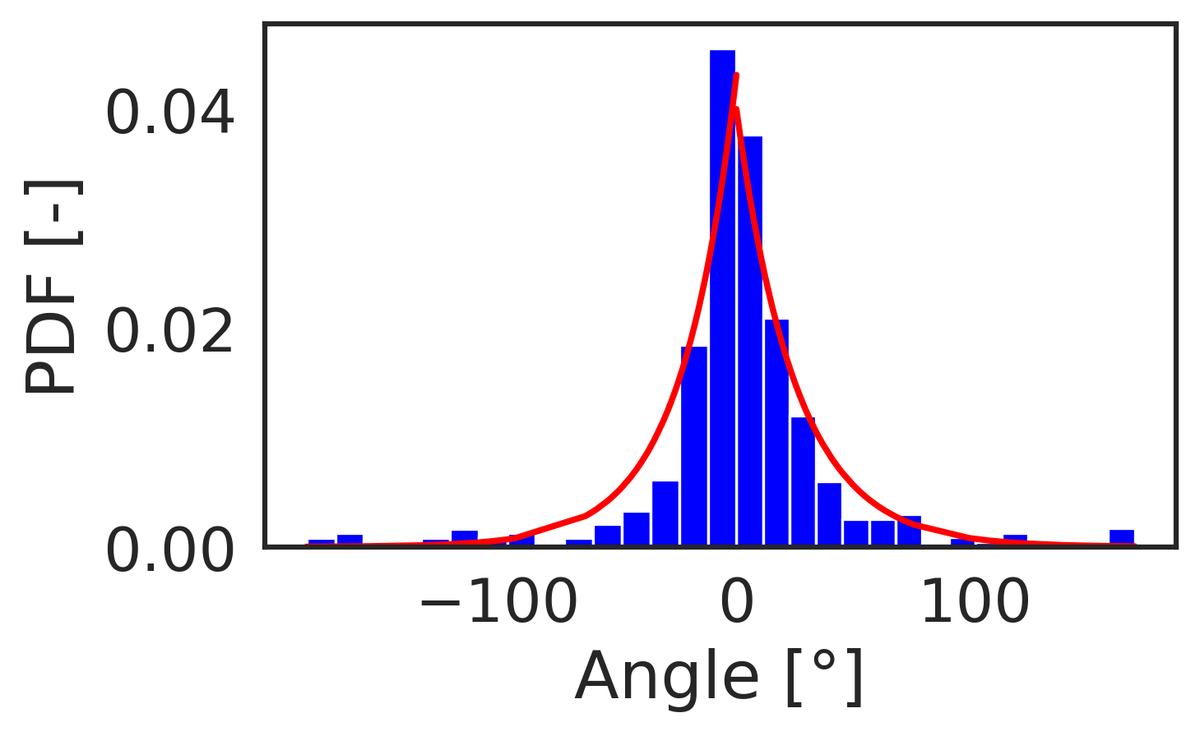}
\caption{}
\label{fig: bio: expoential distribution of steering angle at 146 energy}
\end{subfigure}
\end{minipage}
\hfill
\begin{minipage}[b]{0.3\textwidth}
\begin{subfigure}{\textwidth}
\includegraphics[width=\textwidth]{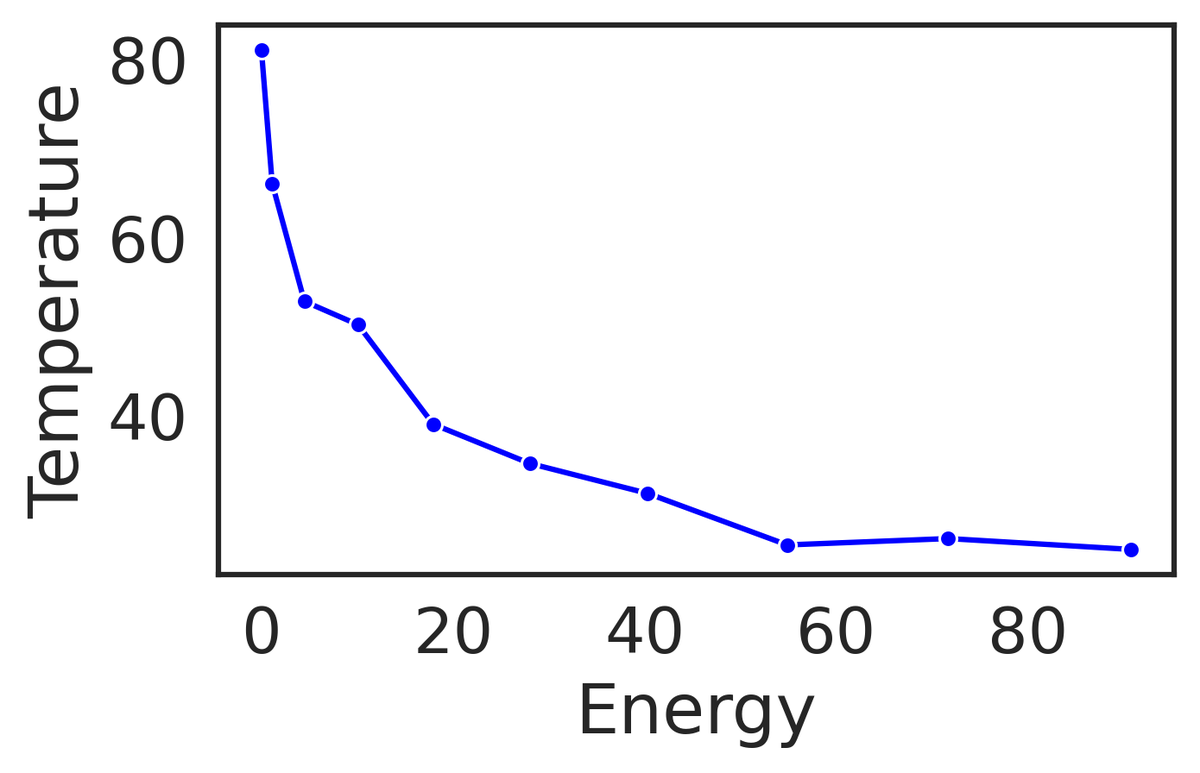}
\caption{}
\label{fig: bio: T of steering angle at different speed}
\end{subfigure}
\end{minipage}
\caption{Steering angle rate distribution at different energy level. (a) 4.5 (ks-statistic: 0.060, p-value: 0.3) (b) 10, (ks-statistic: 0.060, p-value: 0.3)  (c) 18, (ks-statistic: 0.076,  p-value: 0.4) (d) 28, (ks-statistic: 0.076,  p-value: 1)  (e) 40, (ks-statistic: 0.076,  p-value: 1) (f) 54, (ks-statistic: 0.073,  p-value: 1) (g) 71, (ks-statistic: 0.10,  p-value: 1) (h) 90, (ks-statistic: 0.088,  p-value: 1), (i) Temperature of the steering angle rate distribution $p(\theta) = \frac{1}{kT_\theta(v)}e^{-\frac{1}{kT_\theta(v)}\theta}$ at different energy level, where $T$ is a function energy. Intuitively, the steering angle rate range are wider at low energy and low speed.}
\label{fig: bio: steering angle distribution}
\end{figure}

The speed distribution of the ants follows a bimodal exponential form, analogous to the gas and liquid phases in thermodynamic systems (Fig.~\ref{fig: bio: speed distribution different speed}a-h). The phase with low temperature resembles a liquid state and the phase with high temperature resembles a gas-like state. The KS statistic for a single-phase fit is significantly higher than that for a two-phase fit, indicating a statistically better fit with two distinct energy distributions~(Fig.~\ref{fig: bio: speed distribution different speed}j). The proportion of ants associated with the high temperature phase decreases as they enter the narrow road ~(Fig.~\ref{fig: bio: speed distribution different speed}i), meaning that more ants are in the low temperature phase and the low speed expectation in the narrower part of the road. The high temperature component $T_2$ contributes more significantly to the average ant flow. Previous work \cite{Poissonnier2019-uf} showed that flow rate decreases with increasing ant density. Similarly, the temperature component $T_2$ in the speed distribution is inversely correlated with swarm density. The distribution of the steering angle rate at each speed also follow an energy-model (Fig.~\ref{fig: bio: steering angle distribution}a-h). The temperature of the distribution $T_{\theta}$ decreases with the average speed of the ants (Fig.~\ref{fig: bio: steering angle distribution}i), indicating greater steering variability at low speeds and more constrained steering at higher speeds.  

\section{Robotic experiments: apply stochastic mechanism to robotic swarm decisions}
\label{sec: robotic}
\begin{figure}[h!]
    \begin{subfigure}{0.124\textwidth}
    \centering
    \includegraphics[width=\textwidth]{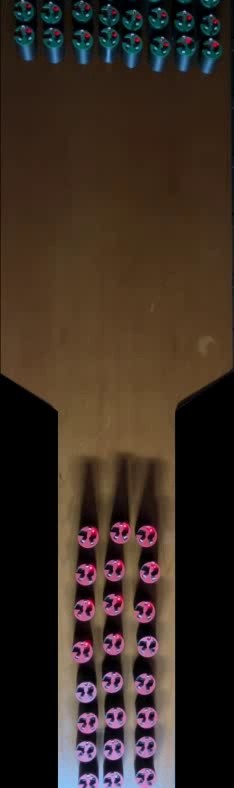}
    \caption{0 s}
    \end{subfigure}
    \hspace{-10pt}
    \begin{subfigure}{0.124\textwidth}
    \includegraphics[width=\textwidth]{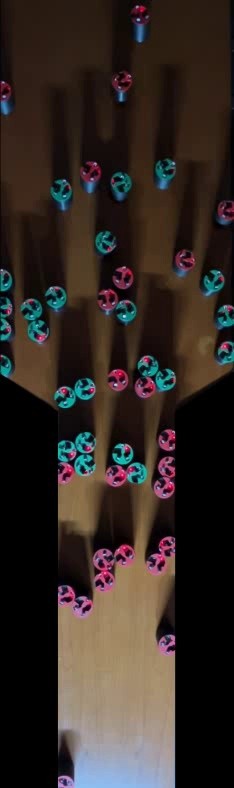}
    \caption{400s}
    \end{subfigure}
    \hspace{-10pt}
    \begin{subfigure}{0.124\textwidth}
    \includegraphics[width=\textwidth]{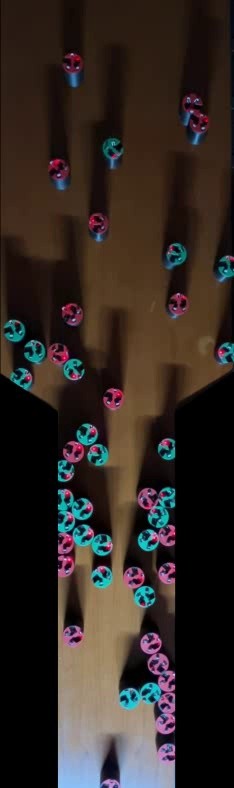}
    \caption{800s}
    \end{subfigure}
    \hspace{-10pt}
    \begin{subfigure}{0.124\textwidth}
    \includegraphics[width=\textwidth]{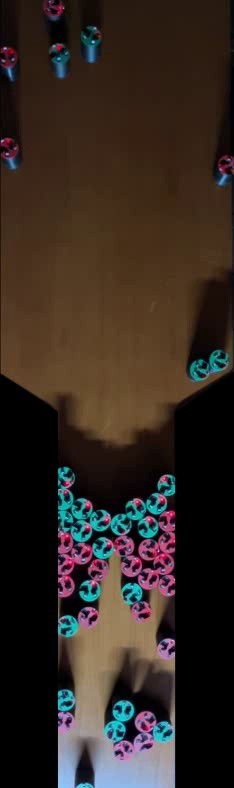}
    \caption{1200s}
    \end{subfigure}
    \hspace{-10pt}
    \begin{subfigure}{0.124\textwidth}
    \includegraphics[width=\textwidth]{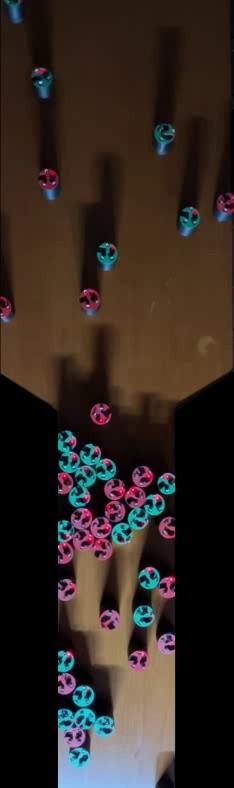}
    \caption{1600s}
    \end{subfigure}
    \hspace{-10pt}
    \begin{subfigure}{0.124\textwidth}
    \includegraphics[width=\textwidth]{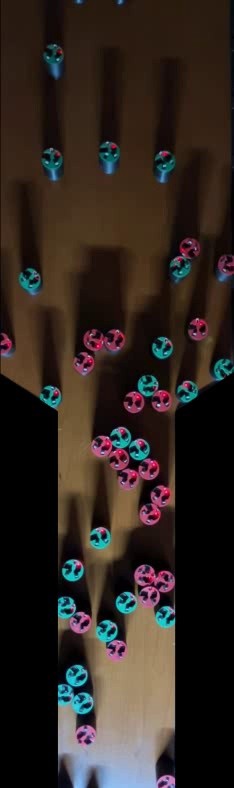}
    \caption{2000s}
    \end{subfigure}
    \hspace{-10pt}
    \begin{subfigure}{0.124\textwidth}
    \includegraphics[width=\textwidth]{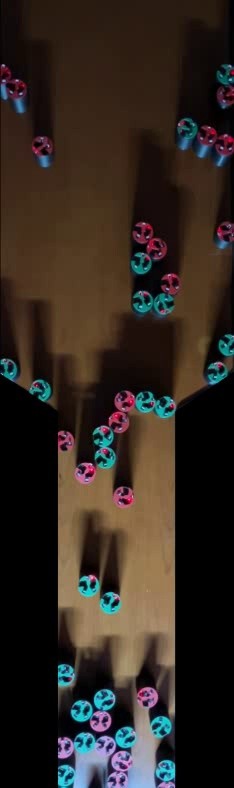}
    \caption{2400s}
    \end{subfigure}
    \hspace{-10pt}
    \begin{subfigure}{0.124\textwidth}
    \includegraphics[width=\textwidth]{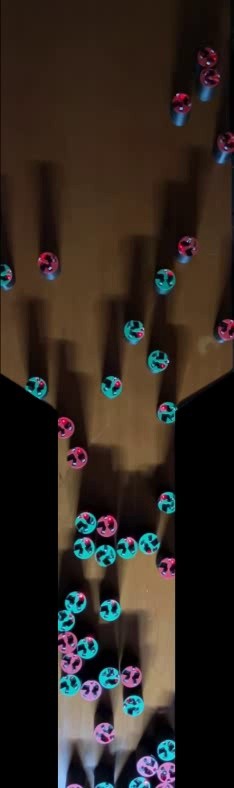}
    \caption{2800s}
    \end{subfigure}
    \hfill
    \begin{subfigure}{0.124\textwidth}
    \includegraphics[width=\textwidth]{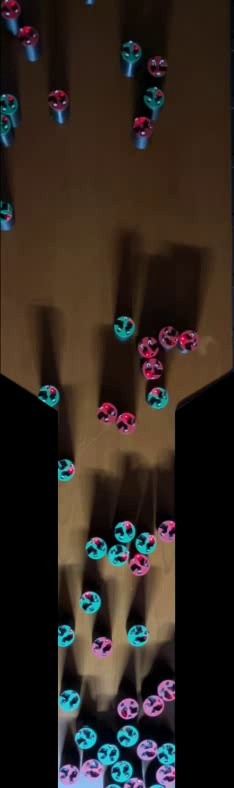}
    \caption{3200s}
    \end{subfigure}
    \hspace{-5pt}
    \begin{subfigure}{0.124\textwidth}
    \includegraphics[width=\textwidth]{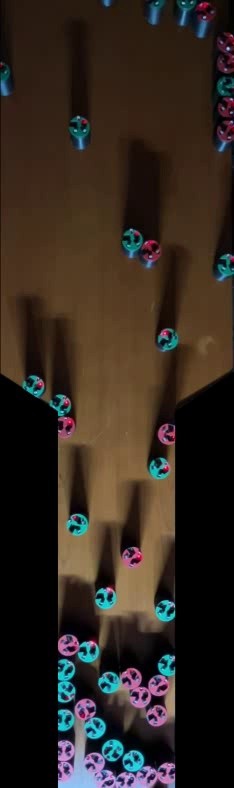}
    \caption{3600s}
    \end{subfigure}
    \hspace{-5.5pt}
    \begin{subfigure}{0.124\textwidth}
    \includegraphics[width=\textwidth]{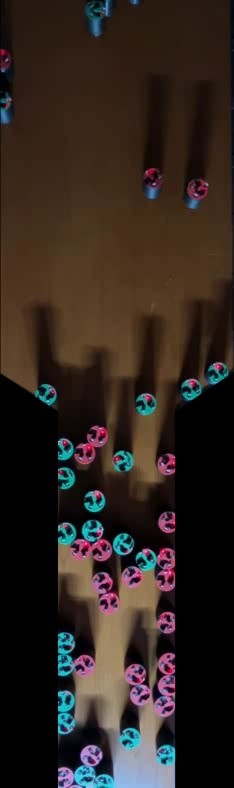}
    \caption{4000s}
    \end{subfigure}
    \hspace{-5.5pt}
    \begin{subfigure}{0.124\textwidth}
    \includegraphics[width=\textwidth]{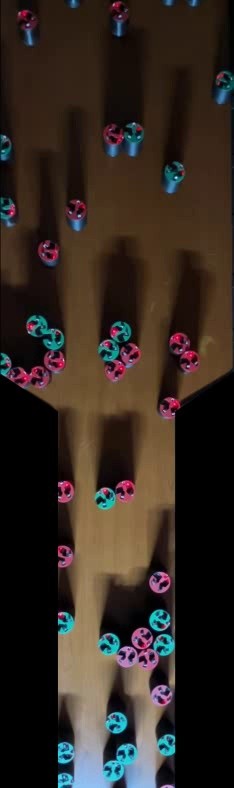}
    \caption{4400s}
    \end{subfigure}
    \hspace{-5.5pt}
    \begin{subfigure}{0.124\textwidth}
    \includegraphics[width=\textwidth]{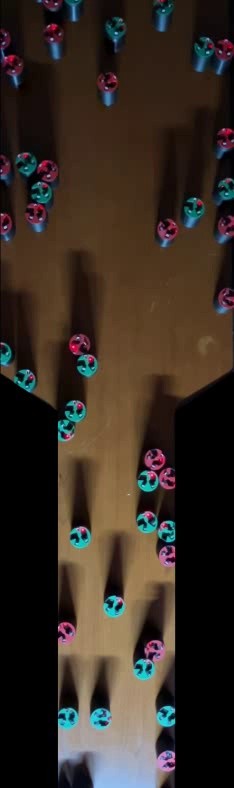}
    \caption{4800s}
    \end{subfigure}
    \hspace{-5.5pt}
    \begin{subfigure}{0.124\textwidth}
    \includegraphics[width=\textwidth]{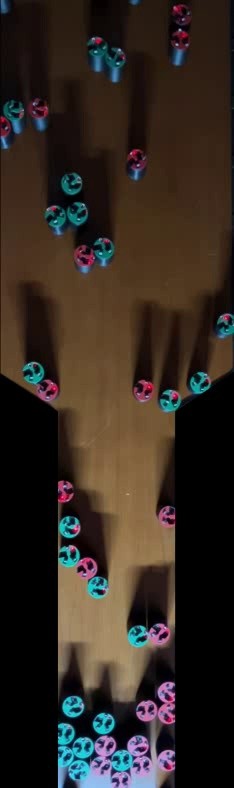}
    \caption{5200s}
    \end{subfigure}
    \hspace{-5.5pt}
    \begin{subfigure}{0.124\textwidth}
    \includegraphics[width=\textwidth]{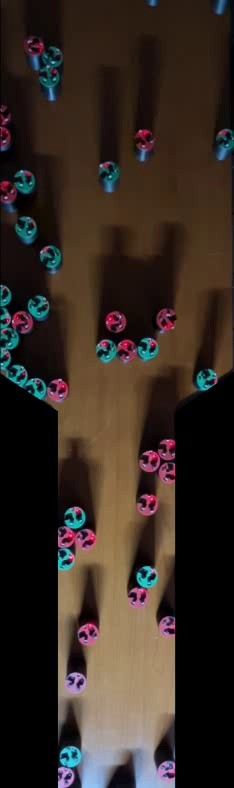}
    \caption{5600s}
    \end{subfigure}
    \hfill
\caption{Long time robotic crossing experiments}
\label{fig: long duration of robot crossing}
\end{figure}
As the second motivation is to replicate the randomness using the discovered energy model from ants to a robotic swarm to achieve scalable and collective behavior with minimum computation. In this framework, actions such as speed and steering angle are modeled as stochastic variables governed by energy-based probability distributions. 

First, we demonstrate that swarm robots governed by the derived principle (Tab.~\ref{tab: speed, angle dist, ant, phys and robot}, Alg.~\ref{alg: narrow}) exhibit flow behavior consistent with that observed in ant swarms. The road configuration (Fig.~\ref{fig: robot experimental settings}a) for robot swarm is the same as for the field experiments. Each robot (Fig.~\ref{fig: field exp settings overview}g), Tab.~\ref{tab:robot hardware config}) is equipped with three photoresistors (Fig.~\ref{fig: robot experimental settings}c) to infer heading direction based on light signals using Alg.\ref{alg: get angle}.
To maintain a spatial density comparable to that of ant swarms, we adjusted the number of robots accordingly (Sec.~\ref{sec: narrow road expeirment configuration}). Unlike ants, which can overlap and exhibit compliance in dense environments, our robots are rigid and cannot overlap. As such, we focus on their ability to traverse the narrow road rather than matching the exact flow rate of the biological system. 

The robot swarm moves bidirectionally through the narrow road, guided by the energy-based control policy (Alg.~\ref{alg: narrow}). There is no communication between each robot, which enables the scalability of the robotic systems. The results (Fig.~\ref{fig: long duration of robot crossing}) demonstrate that robots can pass each other and achieve effective collective navigation, mimicking the cooperative behavior of ants in natural environments.

In the second experiment (Fig.~\ref{fig: simulation, compression demo}), a fixed number of robots was placed in a constrained space, where the upper and lower walls could move vertically under the influence of a constant external force. The robots were controlled using Alg.~\ref{alg: compress experiments}. The temperature of the robot swarm was kept constant in each experiment, while the force applied to the walls varied. As shown in Fig.~\ref{fig: simulation, move without connection 1000 robots, demo}w, the observed pressure–volume relationship is analogous to that of an ideal gas. Swarm robots governed by the energy-based control policy exhibit physical behaviors governed by stochastic mechanisms, consistent with the second law of thermodynamics (Alg.~\ref{alg: compress experiments}). This further demonstrates that the unified principle underlying ant swarms and physical particles also governs the behavior of programmed robotic swarms.

\begin{figure}[htp!]
  \centering
  \begin{subfigure}[t]{0.16\textwidth}
    \centering
    \begin{tikzpicture}
      \node[inner sep=0pt] (img) {\includegraphics[width=\textwidth]{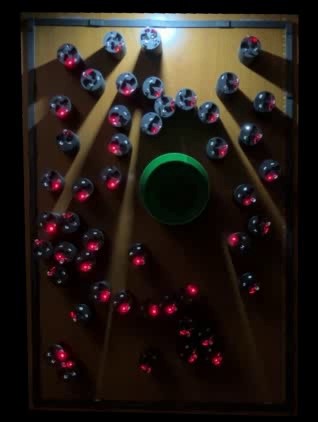}};
      \draw[red, very thick, smooth, tension=0.8]
            (-1.25,0.8) .. controls (-0.4,-0.2) and (0.4,-0.2) .. (1.25,0.8);
      \node[fill=blue!25, rounded corners=2pt,
            text width=1.2cm, align=center, font=\scriptsize,
            anchor=south west] at (-0.7, 0.6)
            {Gas-like};
      \node[fill=blue!25, rounded corners=2pt,
            text width=1.6cm, align=center, font=\scriptsize,
            anchor=north] at (0, -0.8)
            {Liquid-like};
    \end{tikzpicture}
    \caption{0 s}
  \end{subfigure}%
  \hfill %
  \begin{subfigure}[t]{0.16\textwidth}\centering
    \includegraphics[width=\textwidth]{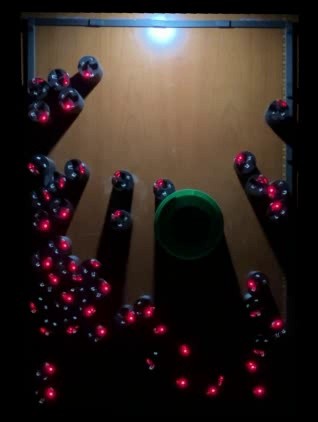}
    \caption{60 s}
  \end{subfigure}\hfill
  \begin{subfigure}[t]{0.16\textwidth}\centering
    \includegraphics[width=\textwidth]{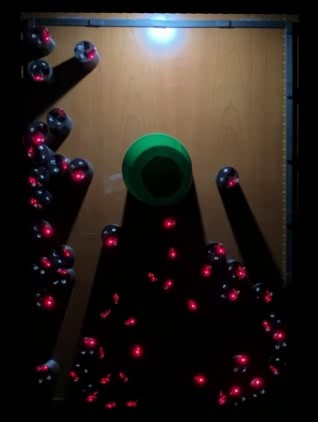}
    \caption{120 s}
  \end{subfigure}\hfill
  \begin{subfigure}[t]{0.16\textwidth}\centering
    \includegraphics[width=\textwidth]{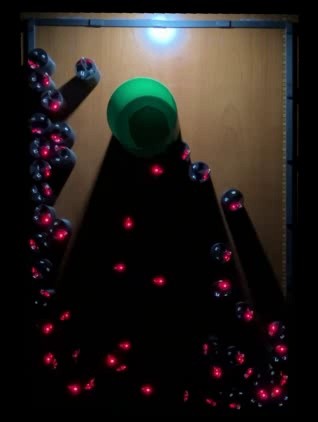}
    \caption{180 s}
  \end{subfigure}\hfill
  \begin{subfigure}[t]{0.16\textwidth}\centering
    \includegraphics[width=\textwidth]{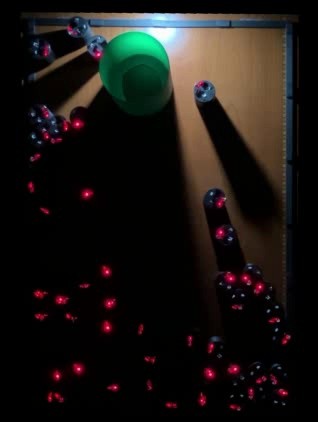}
    \caption{240 s}
  \end{subfigure}\hfill
  \begin{subfigure}[t]{0.16\textwidth}\centering
    \includegraphics[width=\textwidth]{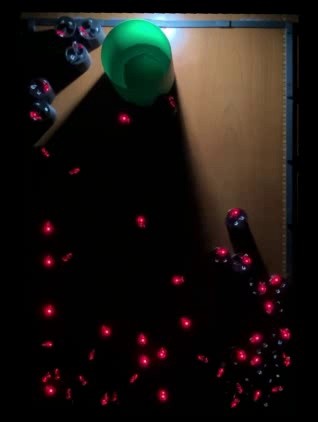}
    \caption{300 s}
  \end{subfigure}

  \caption{Robots move the green object toward the light sources.  
           Robots in illuminated regions (red LEDs visible) behave in a
           \textbf{gas-like phase}; those in darkness remain in a
           \textbf{liquid-like phase}.}
  \label{fig:robot-moving-two-phases}
\end{figure}
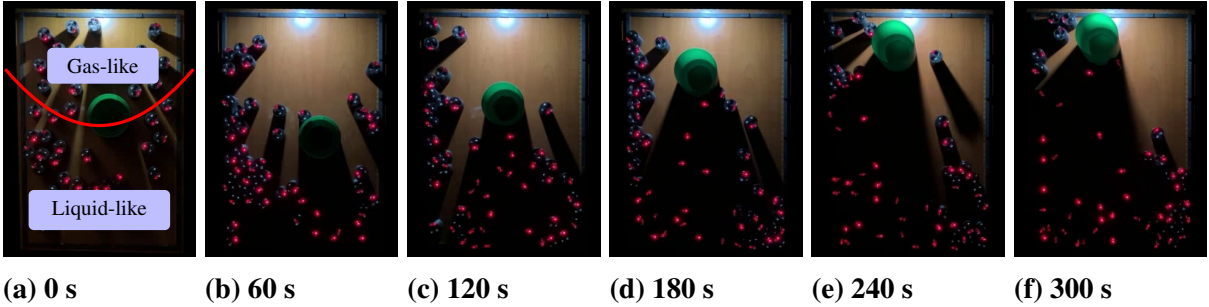

\begin{figure}
    \centering
    \includegraphics[width=0.5\linewidth]{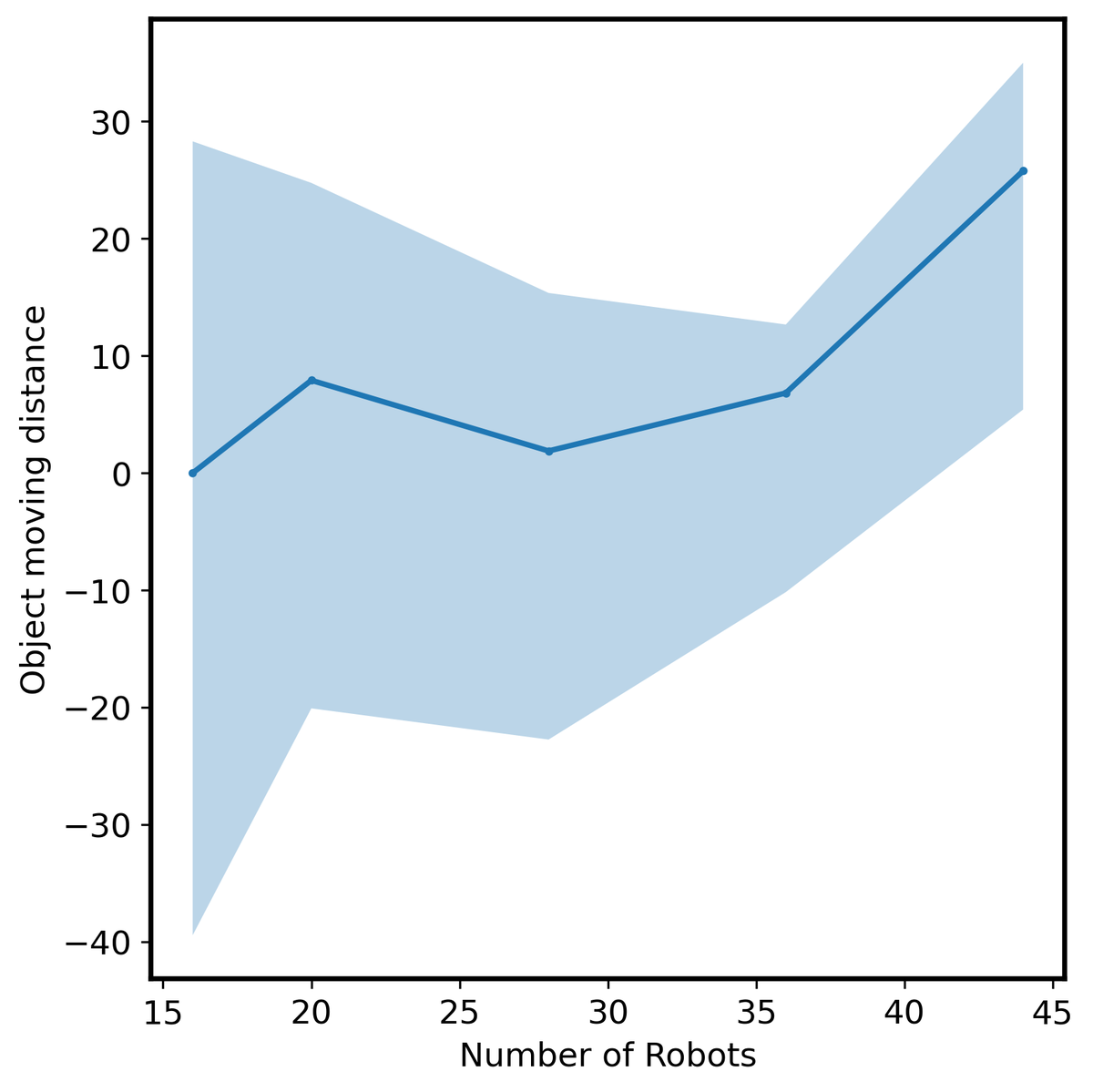}
    \caption{The moving distance of the object(positive direction is towards the lights) increases with number of robots. The variation of object movement is very high and around 0 when the number of robot is small.}
    \label{fig:robot: scalling}
\end{figure}

In the final experiment, we evaluated whether the proposed principle can support intelligent behaviors such as object transport, inspired by how ants move objects. We first show that the robots physically connected to the object successfully moved the object toward the goal (Fig.~\ref{fig: simulation: robot move object without physical connection}) using Alg.~\ref{alg: move with connection}. Next, we disabled the robots’ physical connection to the object. Each robot followed the same control algorithm, requiring minimal computation based on a temperature function derived from the position of the object and the goal (Alg.~\ref{alg: move without connection}). The temperature function depends on a single parameter that switches between high and low values based on the local light level. The light levels are high between the object and the target, whereas the area behind the object remains shadowed. This policy (Alg.~\ref{alg: move without connection}) drives the robot swarm to move the object toward the target (Fig.~\ref{fig:robot-moving-two-phases}). A clear phase separation emerges within the swarm: robots in dark regions exhibit low temperature and behave like a liquid phase, while those between the object and the light source exhibit high temperature, corresponding to a gas-like phase. (Fig.~\ref{fig:robot-moving-two-phases}.a, Fig.~\ref{fig: simulation, move without connection 1000 robots, demo}). This experiment shows that the swarm not only performs intelligent object transport but also retains physical characteristics such as phase separation. The performance of the robotic swarm increases with the number of robots (Fig.~\ref{fig: simulation, move without connection 1000 robots, demo}, Fig.~\ref{fig:robot: scalling}), while the computational load per robots does not increase and is kept minimum.

\begin{figure}[H]
    \centering
    \begin{minipage}[b]{\textwidth}
    \begin{subfigure}{\textwidth}
     \includegraphics[width=0.98\textwidth]{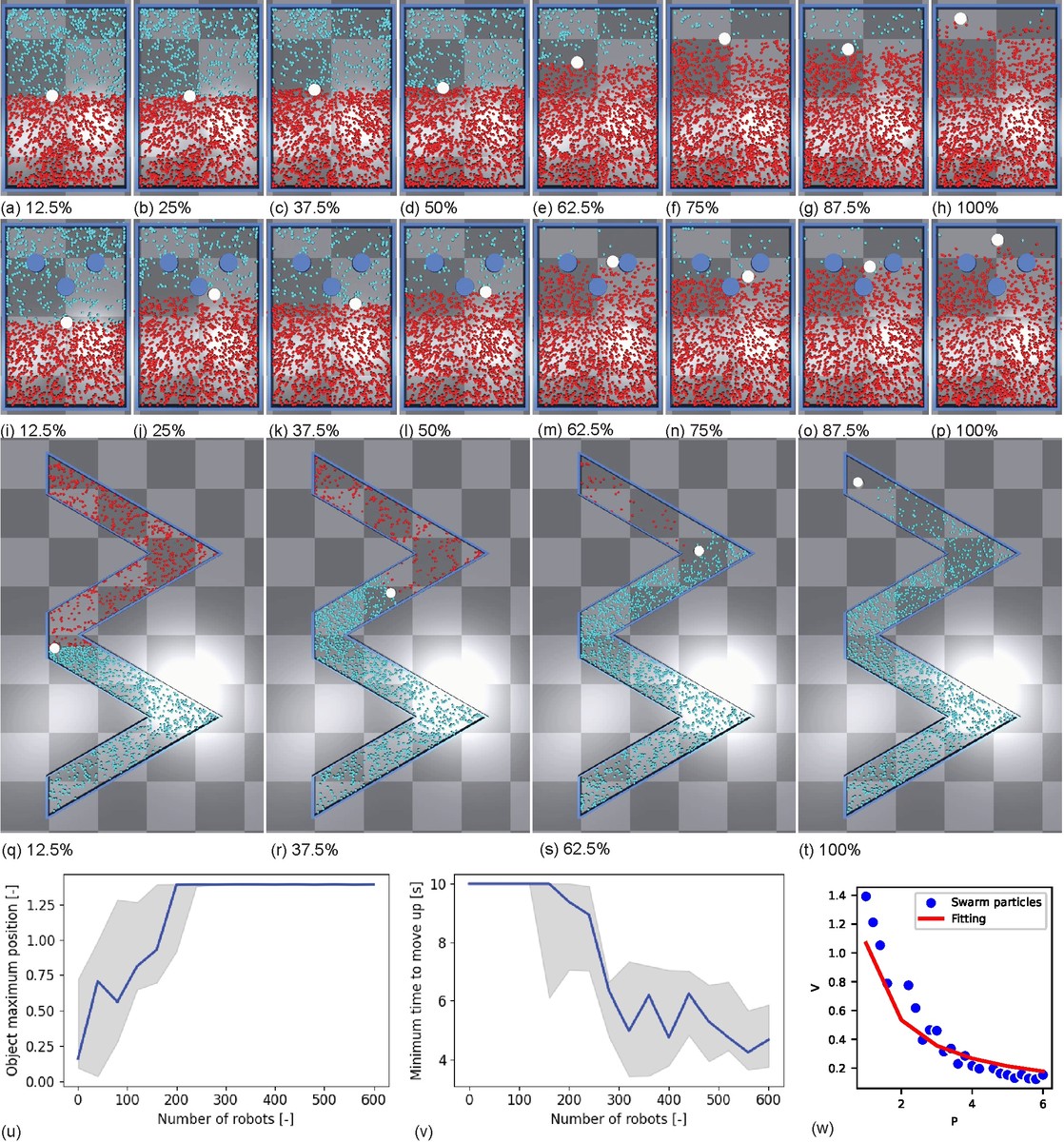}
    \end{subfigure}
    \end{minipage}
    \caption{1200 robots move object without physical connection: (a-h): robot swarm move object without obstacle to the top, (i-p): robot swarm move object with obstacle to the top, (q-t): robot swarm move object zig curve to the top. Capacity quantification of robotic swarms: u) The capacity of robotic swarm quantified by maximum object moving distance, increase with number of robots, (v) The minimum time to move object to desired position, decreases with number of robots. (w) Temperature increases to keep density the same in compression expeirments}
    \label{fig: simulation, move without connection 1000 robots, demo} 
\end{figure}
\section{Disucssion}
According to the statistical mechanism (Eq.~\ref{eq:opt map <f(x_i)> to f(x_i)}, \ref{eq: prob of action under entropy maximization}), the probability density functions of speed and angle represent closed-form solutions to the entropy maximization problem under energy constraints, with expected energy $E = kT$. Intuitively, the individual actions of the ant, such as speed and steering angle rate, are stochastic and not deterministic, but with a common expectation of speed and steering angle rate. 
\begin{table}[h]
    \centering
    \begin{tabular}{|c|c|c|c|c|}
    \hline
                        & Speed                                          & Steering angle rate              & Energy         & Energy   \\
                        &                                                &                                  & (speed)        & (angle)\\\hline
    Ants                & $p(v^2) = \sum_i^2\phi_i \frac{1}{E_1}e^{-\frac{1}{E_1}v^2}$   & $p(\theta) = \frac{1}{E(v)}e^{-\frac{1}{E(v)}\theta}$ & & \\ 
                        & $\sum_i^1\phi_i=1$                             &                                  & $E_i, i\in[0,1]$          & $E(v)$                                \\\hline
    Physical            & $p(v^2) = \frac{1}{E}e^{-\frac{1}{E}v^2}$      & Uniform/Random                   & $E=kT$         &  - \\ 
    particles           &                                                &                                  &                &   \\ \hline 
    Swarm Robots        & $p(v^2) = \frac{1}{E(x)}e^{-\frac{1}{E(x)}v^2}$& Uniform/Random                   & $E(x)$         &  - \\ 
    (move)              &                                                &                                  &                &   \\ \hline 
    Swarm Robots        & $p(v^2) = \frac{1}{E}e^{-\frac{1}{E}v^2}$& $p(\theta) = \frac{1}{E(v)}e^{-\frac{1}{E(v)\theta}}$ & $E$ & $E(v)$ \\ 
    (cross)             &                                                &                                  &                &   \\ \hline 
    \end{tabular}
    \caption{Comparison of distribution of speed and steering angle rate of ants, physical particles and robot swarm. The distribution of speed and angle are the closed form solution of entropy maximization (Eq.~\ref{eq:opt map <f(x_i)> to f(x_i)}). The main differences are what the probability represents and the energy function in the constraints (Eq.~\ref{eq:opt map <f(x_i)> to f(x_i)}). The energy function of speed($v$) related to the density of the ant, the energy function of steer angle($\theta$) of the ants related to the speed. The energy function of physical particles are the system state which can be measured by temperature($T$,$k$ is a constant). The energy function of speed ($v$) of swarm robots in crossing experiments is a constant, and that of steering angle is a function of speed. The energy function of speed ($v$) of swarm robots in moving object experiments is a function of relative position of the object ($x$) measured by the lightness. }
    \label{tab: speed, angle dist, ant, phys and robot}
\end{table}

Table~\ref{tab: speed, angle dist, ant, phys and robot} summarizes the speed and steering angle distributions of ants, robots along with those of the physical particles. In the classical Boltzmann distribution~\cite{boltzmann1877beziehung}, physical particle speed, such as that of atoms or molecules, is governed by a single energy component $E_1$, since the system is characterized by a single equilibrium temperature. In contrast, the speed distribution of \textit{Formica polyctena} exhibits two energy components, $E_1$ and $E_2$, yet retains the same exponential form of the energy distribution. 
Although the direction of motion of the bulk particles is typically random or uniformly distributed, the steering angle rate of the ants also follows an energy-based distribution, with the energy $E = kT$ varying as a function of speed (Fig.~\ref{fig: bio: steering angle distribution}i).
The energy function of speed and the steering angle rate in an ant swarm are biologically programmed, whereas the particles in the bulk of a solid, liquid, or gas are governed by energy exchanges. Although the energy model's arguments are different and have different numbers of energy components, the underlying distributions share a common exponential form, which represents a closed form of entropy maximization. Intuitively, the ants maximize the entropy of the speed and steering angle distributions while maintaining constant average values. This provides a unified statistical explanation for the stochastic behavior observed in both ant swarms and physical particles.
\section{Conclusion}
The energy modeling of The stochastic behavior describes the distribution of individual ant energy in constrained environments and shares the same form as the particle energy distribution in an isolated ideal gas system. This principle enables fully decentralized, robust, and scalable collective robotic systems that do not require communication between units. Robotic swarms governed by this principle exhibit intelligent behaviors, such as object transport, while preserving physical properties such as phase transitions, all in a decentralized fashion. 
This discovery provides a unified theoretical foundation for describing stochastic behavior across ant swarms, physical particles, and robotic swarms. It bridges the domains of statistical physics, biology, and robotics, offering a framework for designing collective systems with minimal computation per robot. 
Looking ahead, the energy function, as a closed-form solution to entropy maximization, enables fully decentralized robotic swarm systems with minimal computation and balanced capability. This approach holds promise for practical applications, ranging from microrobotics in healthcare to large-scale autonomous robotics in warehouses and beyond.

\clearpage
\newpage
\newpage
\section{Methods}
\subsection{Biological experimental settings}
\subsubsection{Biological experimental setup}
\begin{figure}[ht]
    \centering
    \includegraphics[width=0.8\linewidth]{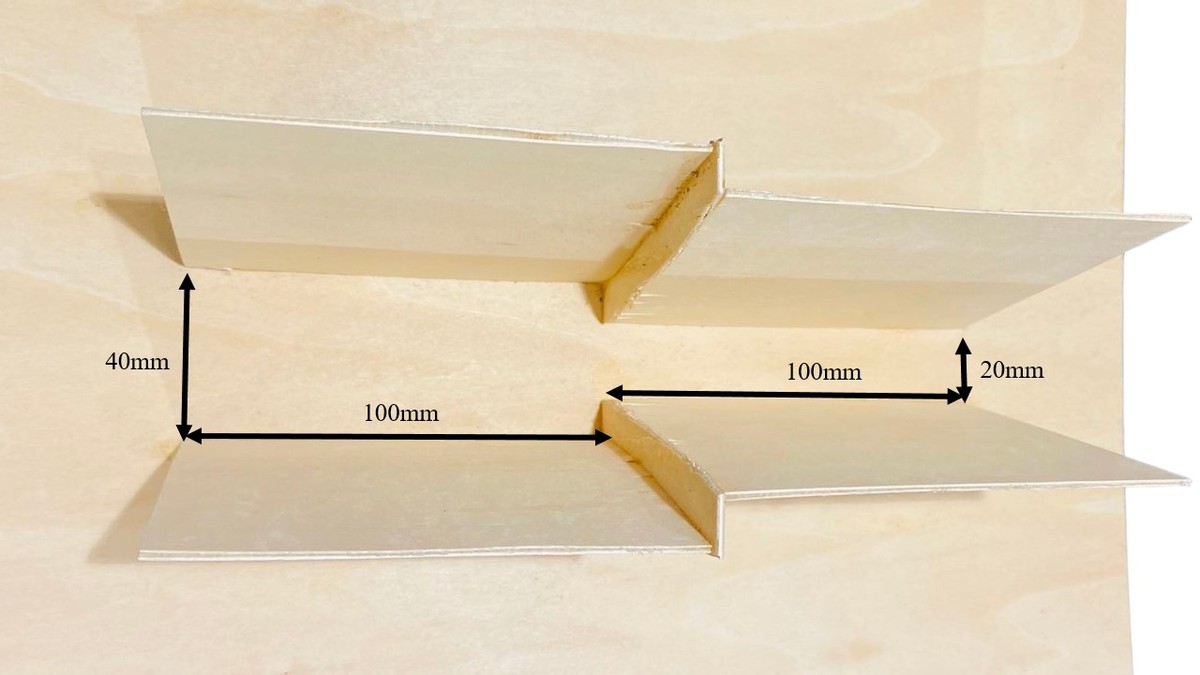}
    \caption{Road dimension}
    \label{fig:road dim}
\end{figure}

The dimensions of three roads are illustrated in Fig.~\ref{fig:road dim}. The narrow straight road has a wide section 40 mm in width and a narrow section 20 mm wide.

The raw image has height of 1920 and width of 1080 and the images are normalized such that 1mm equals to 3.04 pixels.

Temperatures ranged from 72 to 75 \textdegree F and humidity level 43\%-53\%, during the experiments. We compare ant behavior between the upper and lower segments of the same road, ensuring consistent environmental conditions. Therefore, changes within this variation of temperature will not influence the trend shown in this paper.

\subsubsection{Annotation}
\begin{figure}
    \centering
    \includegraphics[width=0.5\textwidth]{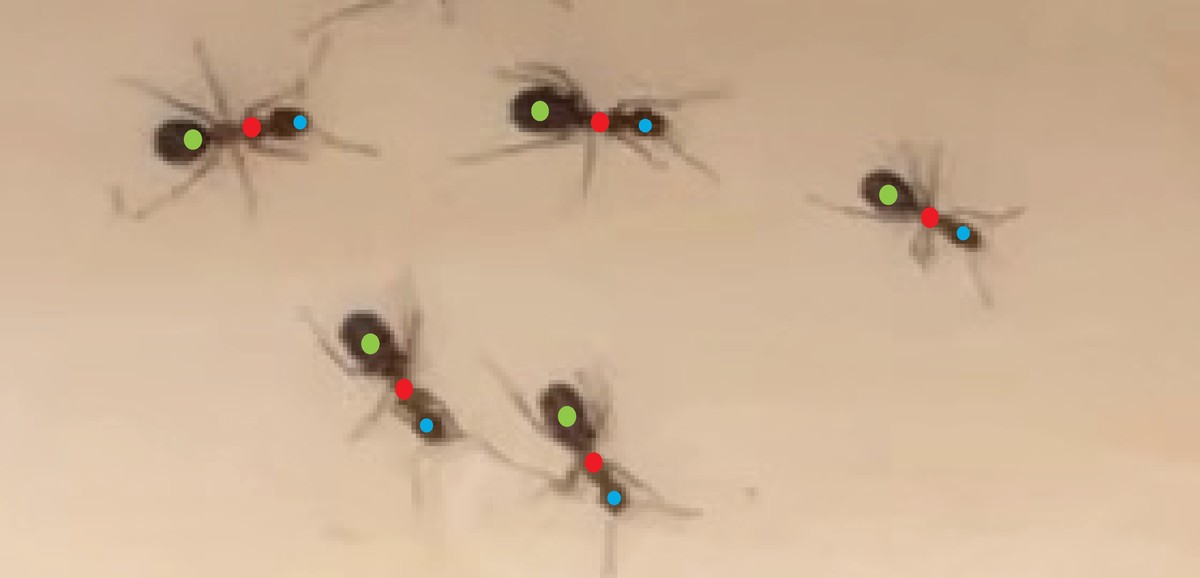}
    \caption{Annotation. Red circle: body, blue circle: head, green circle: tail}
    \label{fig:Annotation of ant}
\end{figure}

For detection of each individual ant, we annotate each ant's pose including three parts: head, body, and tail (seen in Fig.~\ref{fig:Annotation of ant}). We annotated ~413 ants across 31 frames for training and ~57 ants across 4 frames for evaluation. During annotation, the circular center of head, tail, and body parts was annotated as the center. 

\subsection{Video analysis and tracking} \label{sec: method: video analysis and tracking}
\begin{figure}[h!]
    \centering
    \includegraphics[width=1.0\linewidth]{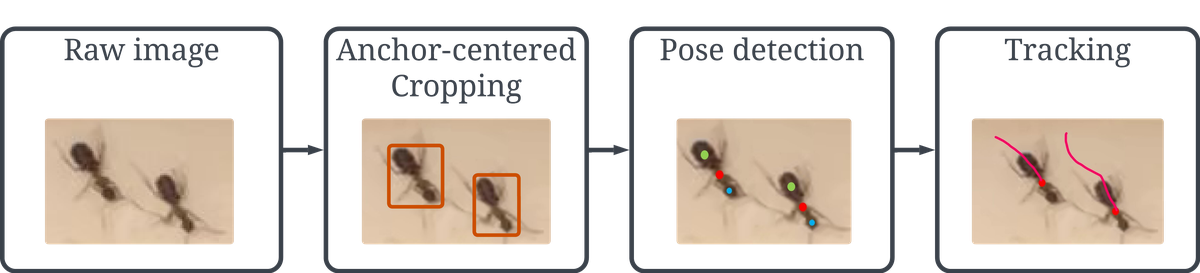}
    \caption{Ant tracking method illustration}
    \label{fig:ant-tracking-flow}
\end{figure}

Tracking was divided into the detection stage and the tracking stage and using the working flow shown in Fig.~\ref{fig:ant-tracking-flow} to track the ants. 

\subsubsection{Detection method}
\begin{figure}
    \centering
    \includegraphics[width=0.5\linewidth]{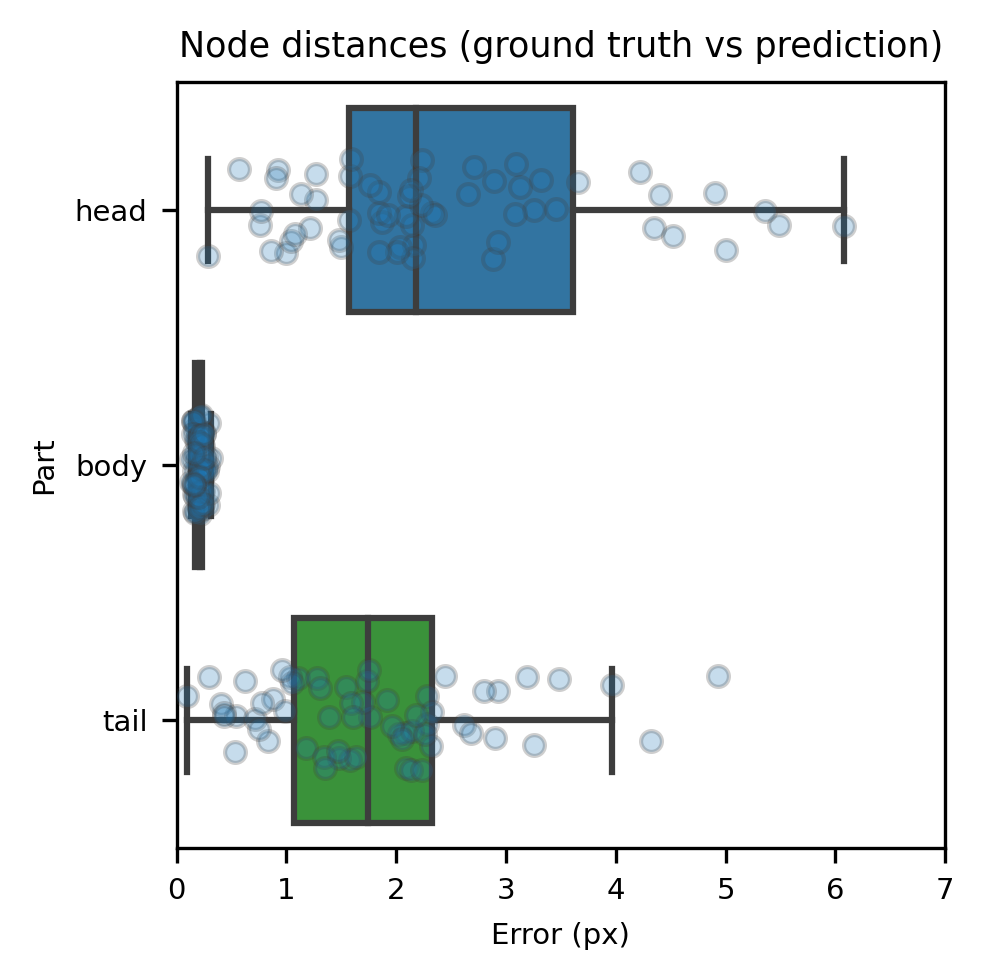}
    \caption{Detection error}
    \label{fig:detection error centroid}
\end{figure}

We used an anchor-based neural network with a U-Net structure \cite{u-net} to detect the peak of each individual ant, then used an anchor-centered cropped around each ant. We used the cropped image of each ant to detect the pose of each ant. The loss of training was the mean squared error of distance for each part of the pose, that is, the head, body, and tail~\cite{Pereira2022-kk}. The precision of the detection of the head, body, and tail is shown in Fig.~\ref{fig:detection error centroid}. 

\subsubsection{Tracking method}

We converted the body, head, and tail to a body-centered bounding box and then calculated the similarity of the bounding box at the current frame and the predicted box of the track using previous frames. The similarity is defined as the feature distance between the detection box in the current frame and the predicted track box. We adopted an adaptive Kalman filter to predict the boundary box of the tracks. In the end, we incorporate the Hungarian Algorithm~\cite{Kuhn1955-jj} to match the bounding box of this frame and previous frames based on similarity. 

\begin{figure}
    \centering
    \includegraphics[width=0.15\linewidth,angle=90]{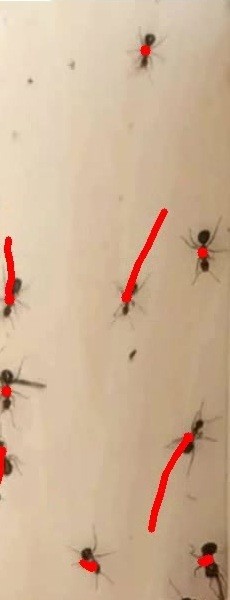}
    \caption{Tracking result demo}
    \label{fig:trackingdemo}
\end{figure}

The MOTA (Multiple Object Tracking Accuracy) is defined as the counts of FP (false positive), FN (false negative) and id-switches (IDSW) normalized over the total number of ground truth (GT) tracks.
\begin{equation}
    \label{eq: MOTA}
    \begin{aligned}
        r_{MOTA} = 1 - \frac{\sum_{t}^{N}(r_{FN_t} + r_{FP_t} + r_{IDSW_t})}{\sum_{t}^{N} r_{GT_t}}
    \end{aligned}
\end{equation}
The accuracy of MOTA is 90.58 \%, the ID swap rate is less than 1.04\%. The tracking results are illustrated in Fig.~\ref{fig:trackingdemo}%

\subsection{Swarm tracking analysis}
\subsubsection{Formulation of energy for the ant}
We revisit the entropy maximization formulation, as shown in Eq.~\ref{eq:opt map <f(x_i)> to f(x_i)} ~\cite{Jaynes1957-ht}. 
\begin{equation} 
\label{eq:opt map <f(x_i)> to f(x_i)}
\begin{aligned}
&\underset{p_i}{\text{max}} 
&& - k \sum_i^n p_i \ln{p_i} \\
& \text{\boldmath{s}.\boldmath{t}.} 
&& \sum_i^{n} p_i = 1 \\
&&&  \| \E(f(x_i)) - E \|=0 \\
\end{aligned}
\end{equation}

where, $x_i \sim p_i(x_i)$. The expectation function $e_i$ in Eq.~\ref{eq:opt map <f(x_i)> to f(x_i)} could be in a form of pressure, density, encoded features, etc. such that $E = \E{(x_i)} = kT$ can be measured. 
The closed form of entropy maximization seen in Eq.~\ref{eq: prob of action under entropy maximization}. 
\begin{equation} \label{eq: prob of action under entropy maximization}
    \begin{aligned}
        p_i & = \frac{1}{E} e^{- \frac{1}{E}x_i} =\frac{1}{kT} e^{- \frac{1}{kT}x_i} \\
    \end{aligned} 
\end{equation}
We summarize that the distribution of the speed and steering angle rate of the ants, and the energy of the physical particles are in an exponential distribution, which are the results of maximization of the entropy but with different constraints of energy (Tab.~\ref{tab: speed, angle dist, ant, phys and robot}). Inspired by such a distribution, our robots swarm have also adopted such distributions in experiments of moving objects and crossing each other in a narrow road.

\subsection{Algorithm}
\subsubsection{Location algorithm}
The localization of robots waswas defined as the heading direction of the robots and the distance towards the light source. The distance towards the light source was calculated by reading the photoresistors. The direction of light source are zero degree for each robot. To let the robot find the zero degree direction, three photoresistors were aligned with the designed angle seen in Fig.~\ref{fig: robot experimental settings}c such that only when $l1$ and $l2$ sensor sense equal value of light and $l3$ sensor detects no light. The zero degree of the robot was aligned by Alg.~\ref{alg: get angle}. 
\begin{algorithm} 
\caption{Align angle function}\label{alg: get angle}
\begin{algorithmic}[1]
\Function{TurnToAngle}{$\theta$}
     \While{Turn one step to zero angle direction}
        \If{$l_1==l2$ and $l_3=0$}
            \State break
        \EndIf
    \EndWhile
\EndFunction
\end{algorithmic}
\end{algorithm}

\subsubsection{Algorithm in narrow road experiments}
The robots executed Alg.~\ref{alg: narrow} for narrow-road experiments both in simulation and robot experiments.
\begin{algorithm} 
\caption{Algorithm for narrow road}\label{alg: narrow}
\begin{algorithmic}[1]
\While{True}
    \State TurnToAngle(0) Alg.~\ref{alg: get angle}
    \State Sample action($\theta_i$,$v_i$) using Eq.~\ref{eq: prob of action under entropy maximization temperature constant},\ref{eq: angle, narrow road}
        \State Robot turn to $\theta_i$ and move with speed $v_i$  
\EndWhile
\end{algorithmic}
\end{algorithm}

In the narrow road experiments, the speed was sampled using the following distribution.
\begin{equation} \label{eq: prob of action under entropy maximization temperature constant}
    \begin{aligned}
        v_i \sim p_i(v_i) & = \frac{1}{kT} e^{- \frac{1}{2kT}v_i^2}
    \end{aligned} 
\end{equation}
The direction of the robot was sampled by normal distribution with the norm of zero degrees for the robot moving down and 180 degrees for the robot moving up. 
\begin{equation} \label{eq: angle, narrow road}
      \theta_{i} \sim
    \begin{cases}
      N(0,180) & \text{if $i$ robot goes up}\\
      N(0,180) & \text{if $i$ robot goes down}
    \end{cases}   
\end{equation}

\subsubsection{Compress experiments}
The robots execute Alg.~\ref{alg: compress experiments} for experiments in narrow road in simulation.
\begin{algorithm} 
\caption{Algorithm for compress experiments}\label{alg: compress experiments}
\begin{algorithmic}[1]
\State Decide the Temperature $T$
\While{True}
    \State TurnToAngle(0) Alg.~\ref{alg: get angle}
    \State Sample action($\theta_i$,$v_i$) using $p_i(v_i)= \frac{1}{kT} e^{- \frac{1}{2kT}v_i^2}$, $\theta \sim U(-180,180)$
        \State Robot turn to $\theta_i$ and move with speed $v_i$  
\EndWhile
\end{algorithmic}
\end{algorithm}

\subsubsection{Moving object with connection}
The robots executed Alg.~\ref{alg: move with connection} for experiments of moving object with connection in simulation.
\begin{algorithm}[!h] 
\caption{Algorithm for moving object with connection}\label{alg: move with connection}
\begin{algorithmic}[1]
\State Decide the Temperature $T$
\While{True}
    \State TurnToAngle(0) Alg.~\ref{alg: get angle}
    \State Sample action($\theta_i$,$v_i$) using $v_i \sim p_i(v_i)= \frac{1}{kT} e^{- \frac{1}{2kT}v_i^2}$, $\theta_i \sim U(-90,90)$
    \State Robot turn to $\theta_i$ and move with speed $v_i$ 
\EndWhile
\end{algorithmic}
\end{algorithm}

\subsubsection{Moving object without connection}
The robots execute Alg.~\ref{alg: move without connection} for experiments in narrow road in simulation.
\begin{algorithm} 
\caption{Algorithm for moving object without connection}\label{alg: move without connection}
\begin{algorithmic}[1]
\State Decide the Temperature $T$
\While{True}
    \State TurnToAngle(0) Alg.~\ref{alg: get angle}
    \State Sample action($\theta_i$,$v_i$) using Eq.~\ref{eq: prob of speed moving object},\ref{eq: temperature dependence in moving object},\ref{eq: angle moving object without connection}
        \State Robot turn to $\theta_i$ and move with speed $v_i$  
\EndWhile
\end{algorithmic}
\end{algorithm}
In the moving object experiments, the speed was sampled using Eq.~\ref{eq: prob of speed moving object}. 
\begin{equation} \label{eq: prob of speed moving object}
    \begin{aligned}
        v_i \sim p_i(v_i) & = \frac{1}{kT_i} e^{- \frac{1}{2kT_i}v_i^2} \\
    \end{aligned} 
\end{equation}
The temperature $T_i$ is a function of lightness $l$ and thread $l_{thread}$ (Eq.~\ref{eq: temperature dependence in moving object}). The steering angle of robots are uniformed sampled as Eq.~\ref{eq: angle moving object without connection}. 
\begin{equation} \label{eq: temperature dependence in moving object}
      T_{i} =
    \begin{cases}
      T_{liquid} & \text{if $l > l_{thread}$}\\
      T_{gas} & \text{if $l > l_{thread}$}
    \end{cases}   
\end{equation}
\begin{equation} \label{eq: angle moving object without connection}
    \theta \sim U(-180,180)
\end{equation}

\subsection{Simulation}
\subsubsection{Narrow road experiment} \label{sec: narrow road expeirment configuration}
We used the ball-shaped robots in the simulation. To keep the spatial distribution of the simulation of the ants and the size of the road proportional to the biological experiments. We keep the occupation ratio the same which is defined by the area size of the ant and the area size of the road.  
On the road shown in Fig.~\ref{fig:road dim}, the ant areas without overlap can have 448 ants to cover the areas, with Ant dimension estimated as 100(length of the road)/14(number of ants in column)=7.14 mm in length and around 20(width of the road)/8(number of ants in row) = 2.5mm in width, we estimated 80 ants on average on the road giving body-to-space ratio of 0.17857. %
We keep the occupation ratio (the area of the ant/area of the road) constant, which gives ball size 33mm in diameters and road length in simulation is 381.

\begin{figure}[h!]
\centering
\begin{minipage}{0.16\textwidth}
    \begin{subfigure}{\textwidth}
    \centering
    \includegraphics[width=\textwidth]{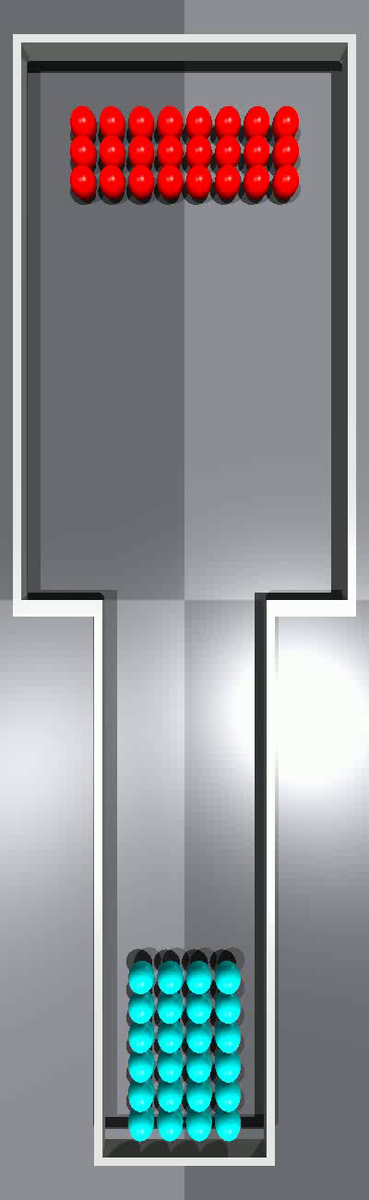}
    \caption{0 s}
    \end{subfigure}
\end{minipage}
\hfill
\begin{minipage}{0.16\textwidth}
    \begin{subfigure}{\textwidth}
    \includegraphics[width=\textwidth]{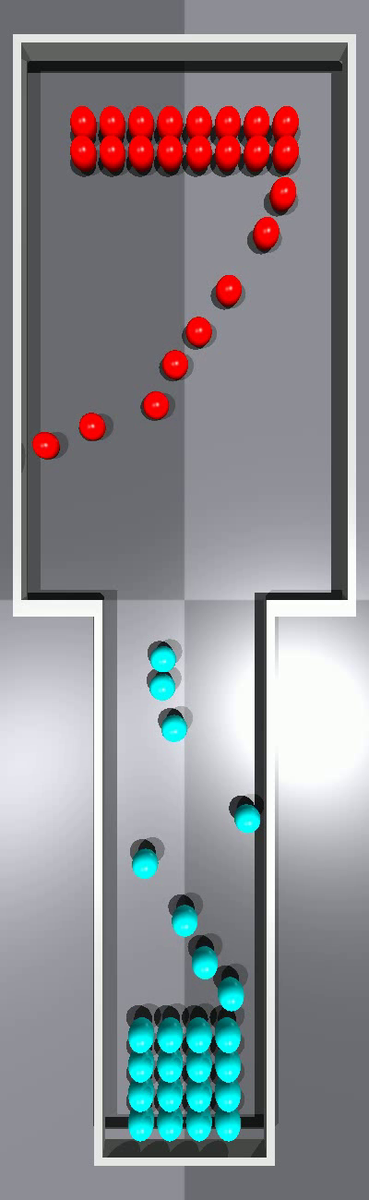}
    \caption{200s}
    \end{subfigure}
\end{minipage}
\hfill
\begin{minipage}{0.16\textwidth}
    \begin{subfigure}{\textwidth}
    \includegraphics[width=\textwidth]{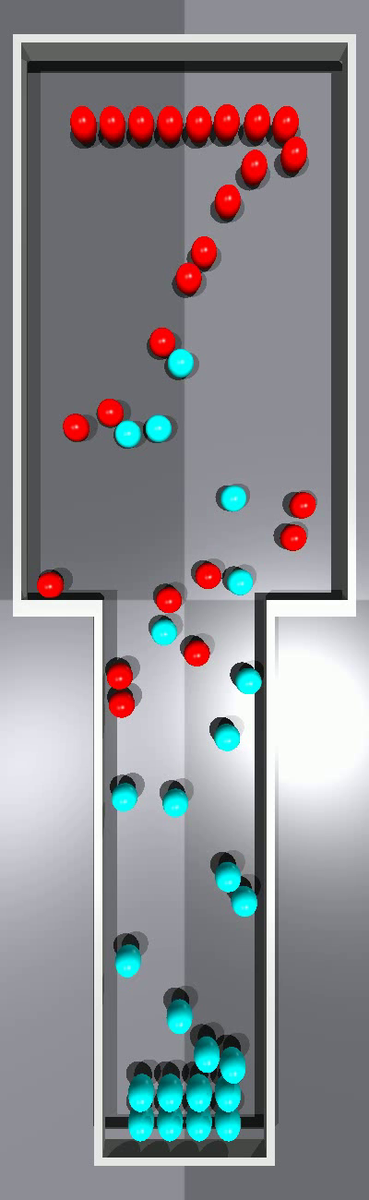}
    \caption{400s}
    \end{subfigure}
\end{minipage}
\hfill
\begin{minipage}{0.16\textwidth}
    \begin{subfigure}{\textwidth}
    \includegraphics[width=\textwidth]{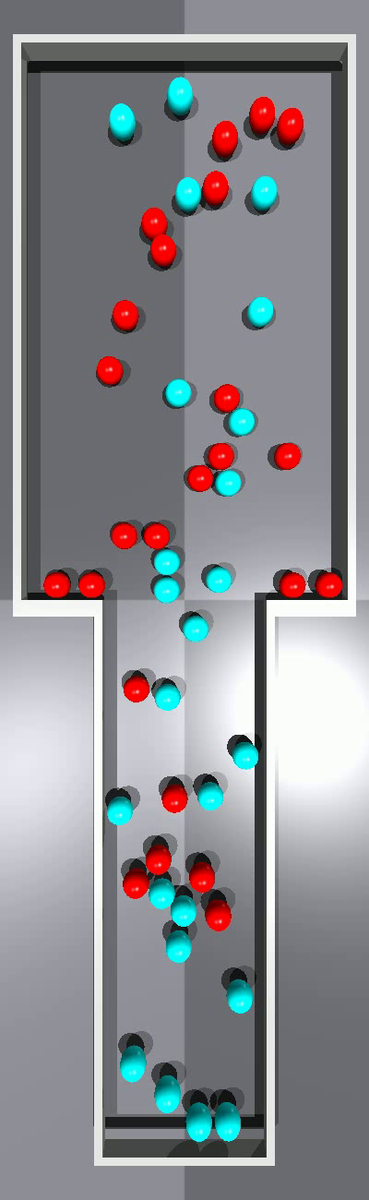}
    \caption{600s}
    \end{subfigure}
\end{minipage}
\hfill
\begin{minipage}{0.16\textwidth}
    \begin{subfigure}{\textwidth}
    \includegraphics[width=\textwidth]{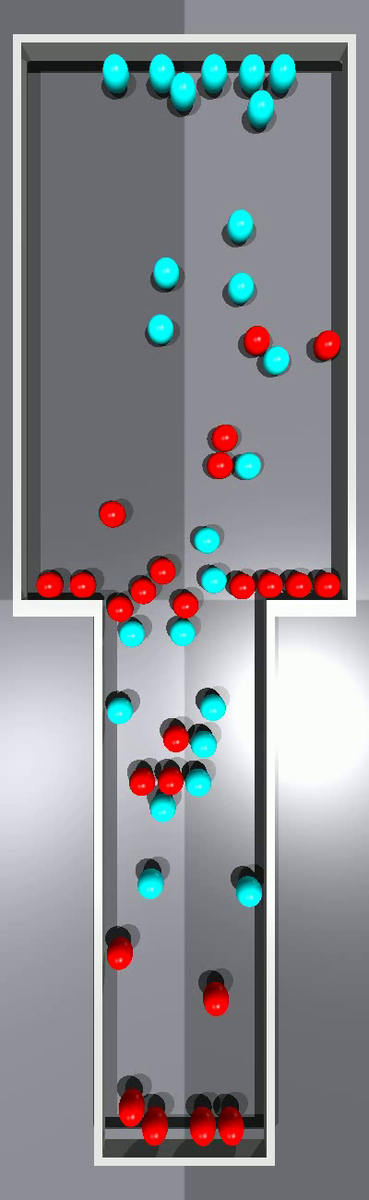}
    \caption{800s}
    \end{subfigure}
\end{minipage}
\hfill
\begin{minipage}{0.16\textwidth}
    \begin{subfigure}{\textwidth}
    \includegraphics[width=\textwidth]{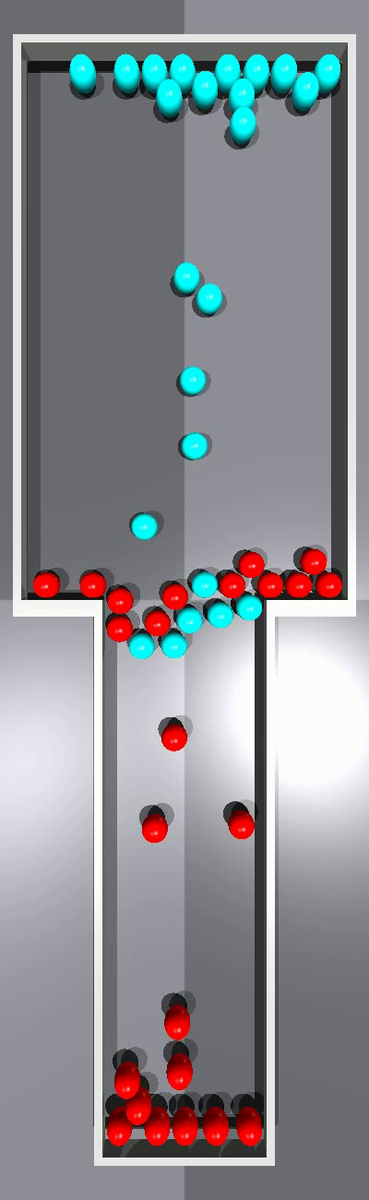}
    \caption{1200s}
    \end{subfigure}
\end{minipage}
\hfill
\begin{minipage}{0.16\textwidth}
    \begin{subfigure}{\textwidth}
    \includegraphics[width=\textwidth]{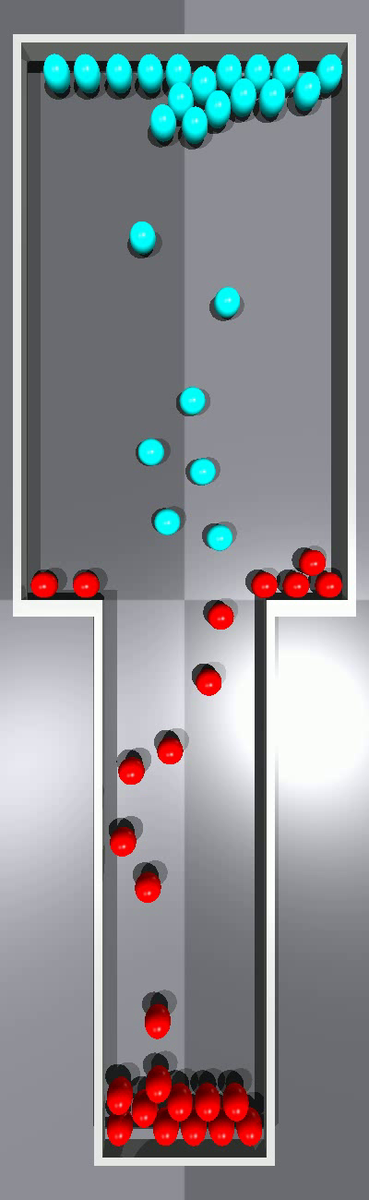}
    \caption{1200s}
    \end{subfigure}
\end{minipage}
\hfill
\begin{minipage}{0.16\textwidth}
    \begin{subfigure}{\textwidth}
    \includegraphics[width=\textwidth]{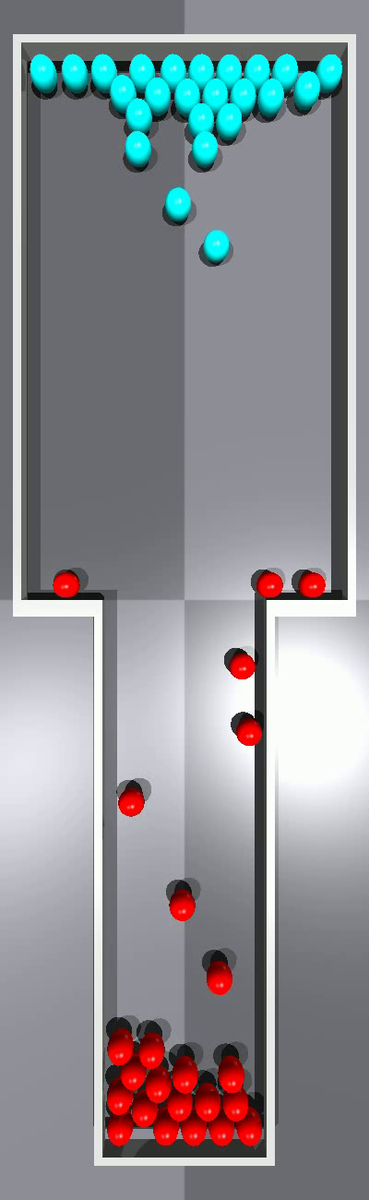}
    \caption{1200s}
    \end{subfigure}
\end{minipage}
\hfill
\begin{minipage}{0.16\textwidth}
    \begin{subfigure}{\textwidth}
    \includegraphics[width=\textwidth]{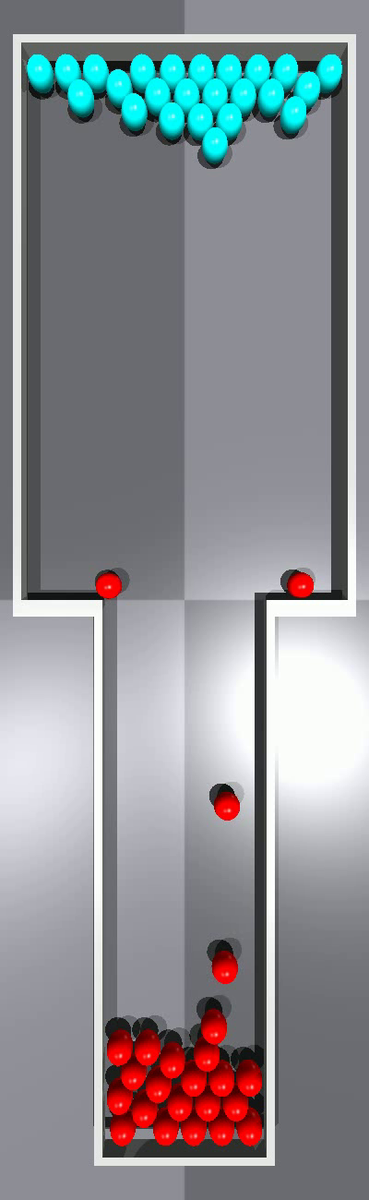}
    \caption{1200s}
    \end{subfigure}
\end{minipage}
\hfill
\hfill
\hfill
\hfill
\hfill
\hfill
\hfill
\hfill
\hfill
\label{fig: simulation, robot crossing congestion}
\caption{Robots crossing congestion}
\end{figure}

\begin{figure}[h!]
\centering
    \begin{subfigure}{0.16\textwidth}
    \centering
    \includegraphics[width=\textwidth]{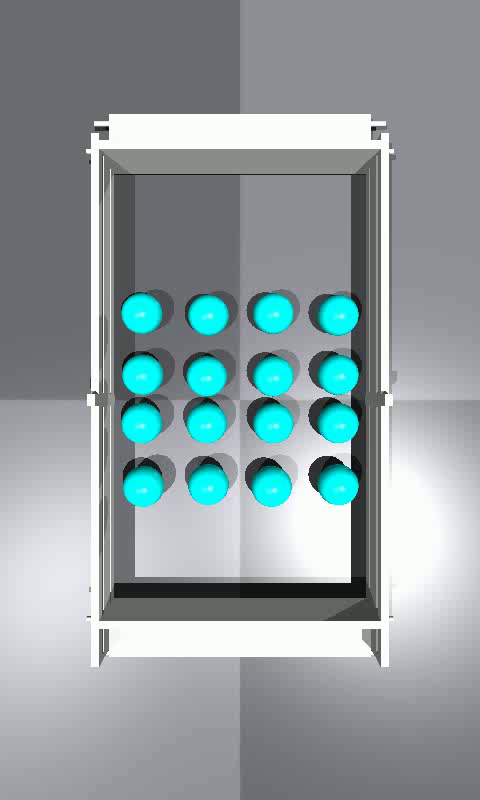}
    \caption{0 s}
    \end{subfigure}
\hfill
    \begin{subfigure}{0.16\textwidth}
    \includegraphics[width=\textwidth]{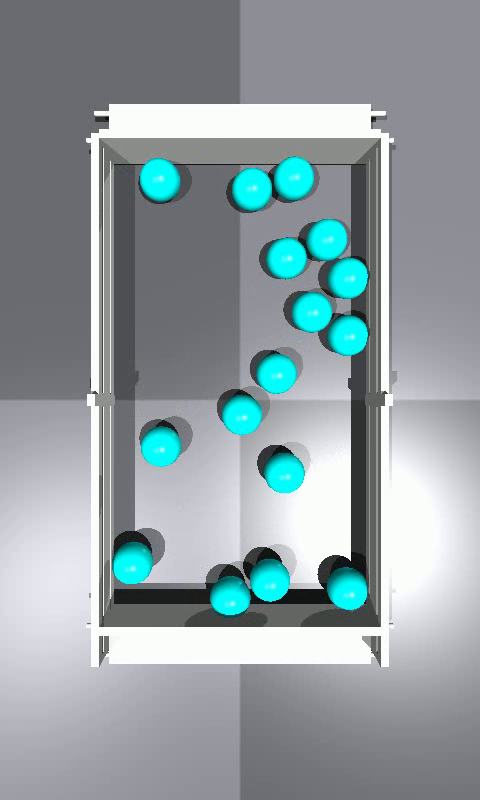}
    \caption{200s}
    \end{subfigure}
\hfill
    \begin{subfigure}{0.16\textwidth}
    \includegraphics[width=\textwidth]{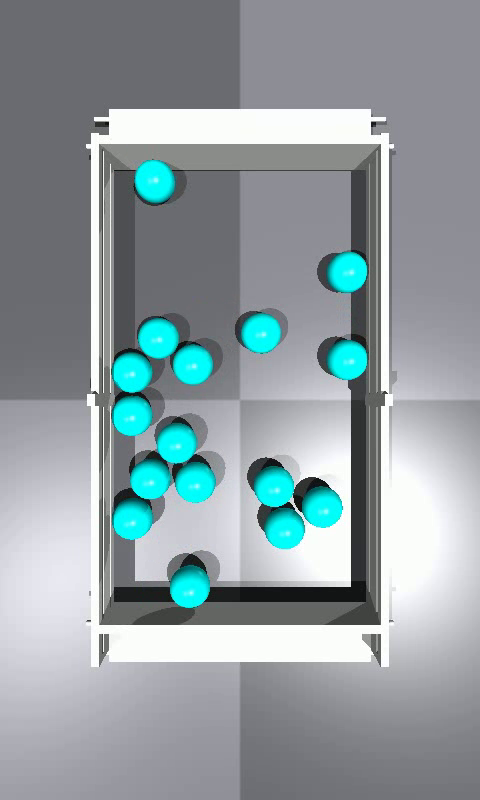}
    \caption{400s}
    \end{subfigure}
\hfill
    \begin{subfigure}{0.16\textwidth}
    \includegraphics[width=\textwidth]{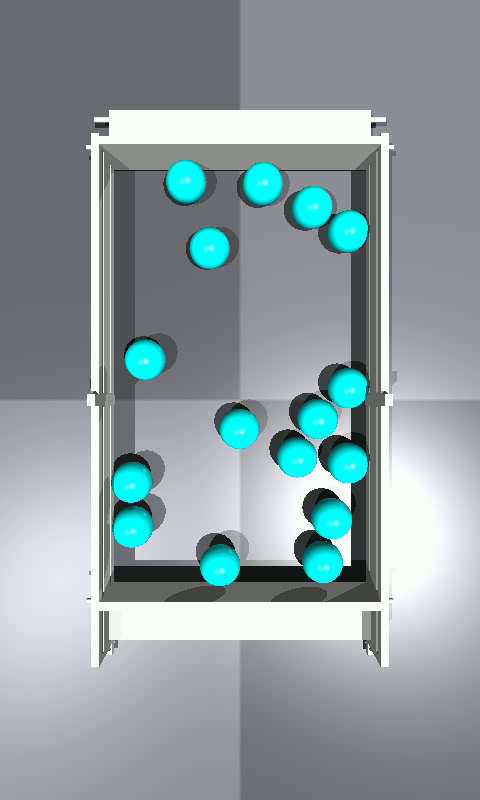}
    \caption{600s}
    \end{subfigure}
\hfill
    \begin{subfigure}{0.16\textwidth}
    \includegraphics[width=\textwidth]{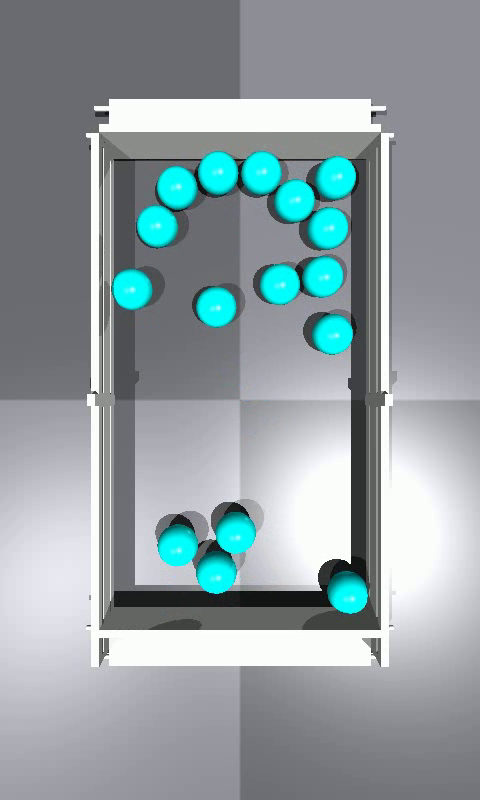}
    \caption{800s}
    \end{subfigure}
\hfill
    \begin{subfigure}{0.16\textwidth}
    \includegraphics[width=\textwidth]{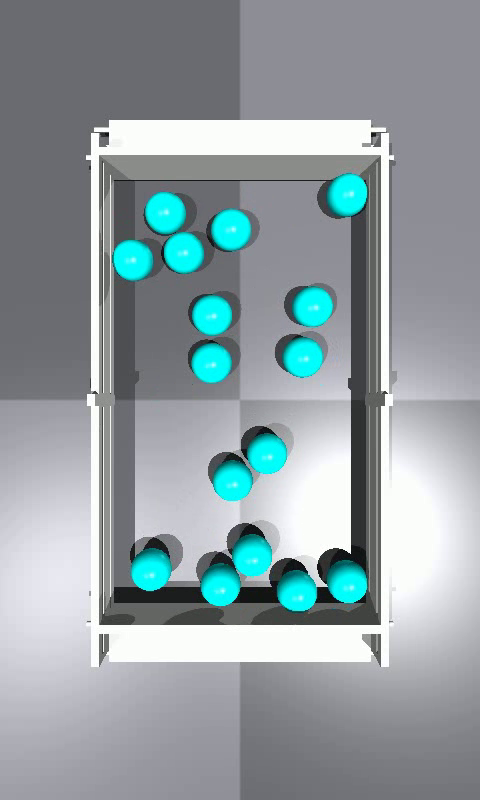}
    \caption{1200s}
    \end{subfigure}
\hfill
    \begin{subfigure}{0.16\textwidth}
    \includegraphics[width=\textwidth]{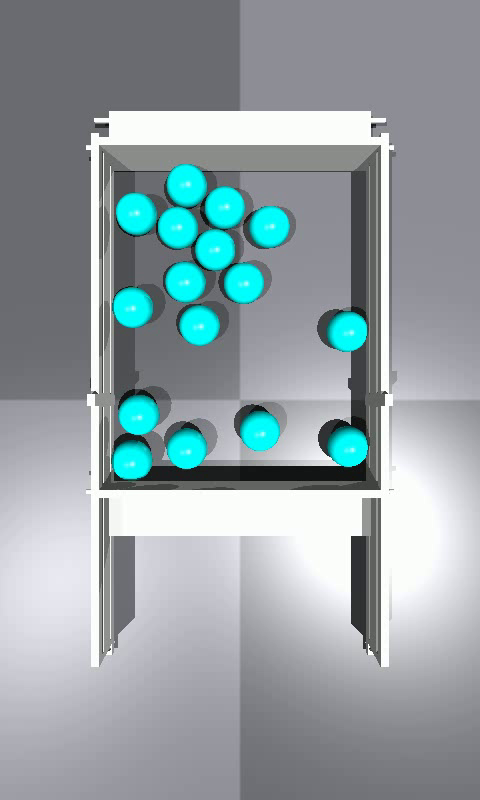}
    \caption{1200s}
    \end{subfigure}
\hfill
    \begin{subfigure}{0.16\textwidth}
    \includegraphics[width=\textwidth]{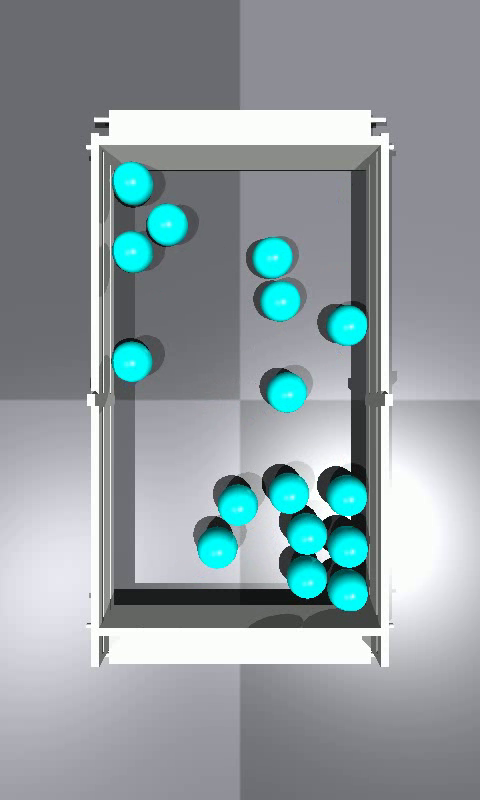}
    \caption{1200s}
    \end{subfigure}
\hfill
    \begin{subfigure}{0.16\textwidth}
    \includegraphics[width=\textwidth]{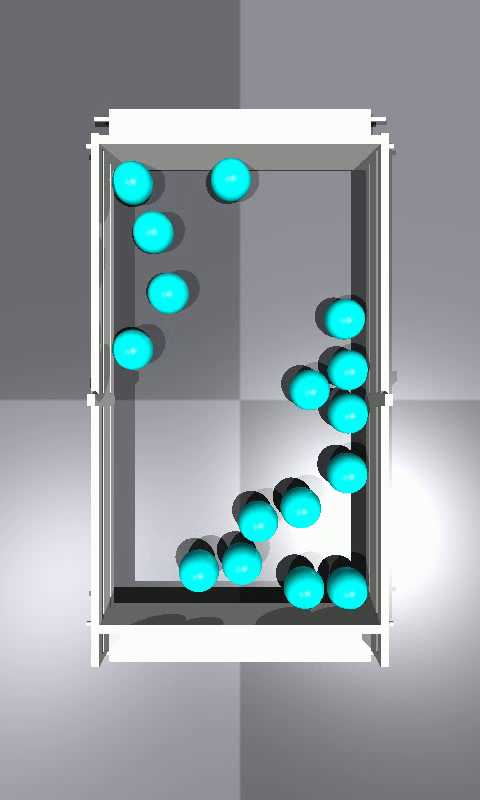}
    \caption{1200s}
    \end{subfigure}
\hfill
\hfill
\hfill
\hfill
\hfill
\hfill
\hfill
\hfill
\hfill
\caption{Compression}
\label{fig: simulation, compression demo}
\end{figure}

\subsubsection{Move object with connection}
\begin{figure}[hbt!] 
\centering
\begin{minipage}{0.16\textwidth}
    \begin{subfigure}{\textwidth}
    \centering
    \includegraphics[width=\textwidth]{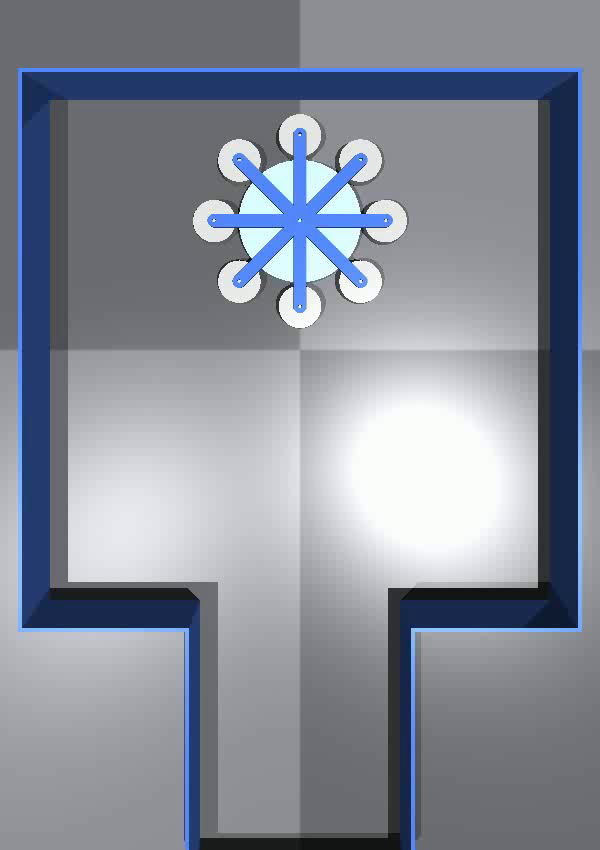}
    \caption{0 s}
    \end{subfigure}
\end{minipage}
\hfill
\begin{minipage}{0.16\textwidth}
    \begin{subfigure}{\textwidth}
    \includegraphics[width=\textwidth]{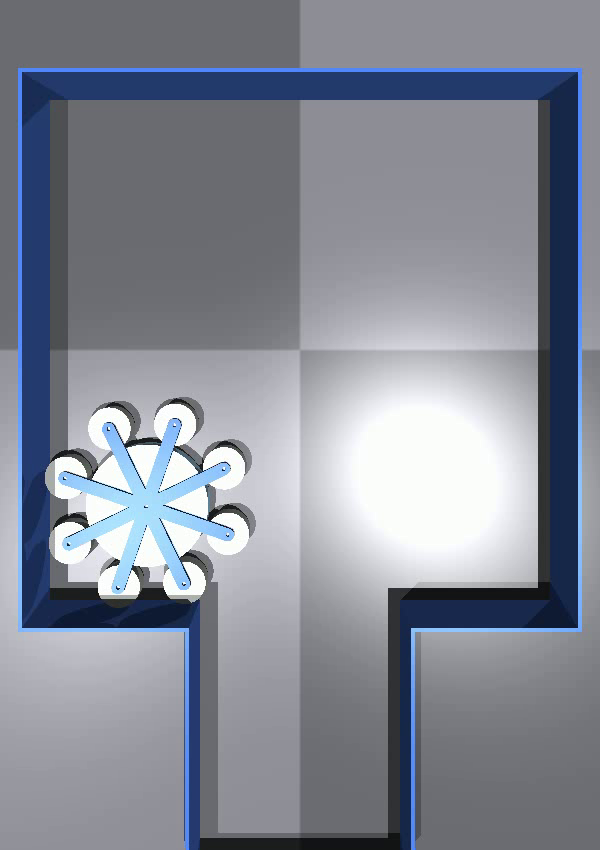}
    \caption{50s}
    \end{subfigure}
\end{minipage}
\hfill
\begin{minipage}{0.16\textwidth}
    \begin{subfigure}{\textwidth}
    \includegraphics[width=\textwidth]{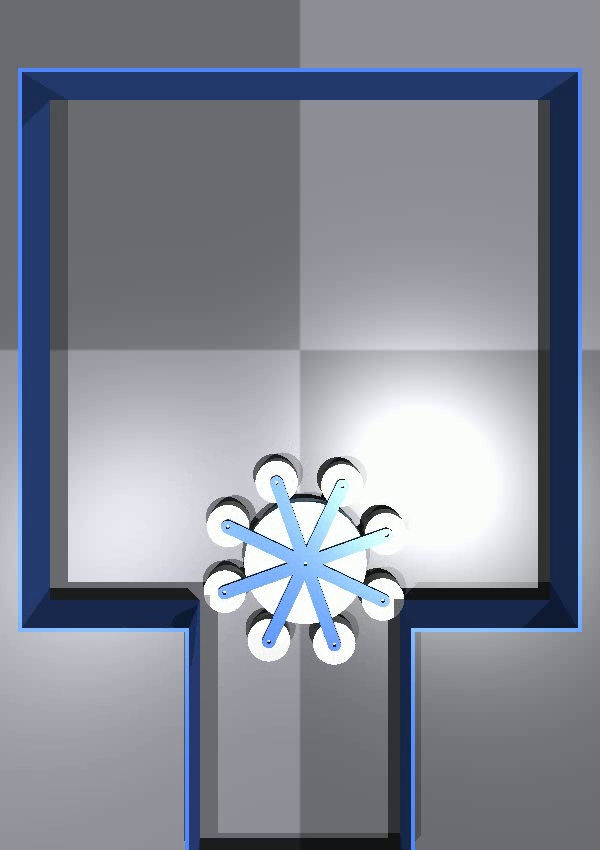}
    \caption{100s}
    \end{subfigure}
\end{minipage}
\hfill
\begin{minipage}{0.16\textwidth}
    \begin{subfigure}{\textwidth}
    \includegraphics[width=\textwidth]{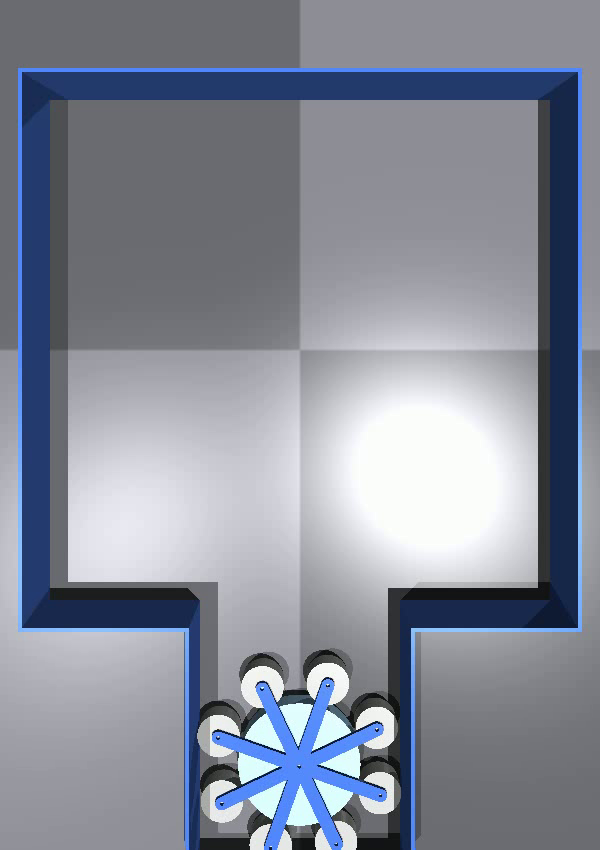}
    \caption{150s}
    \end{subfigure}
\end{minipage}
\hfill
\begin{minipage}{0.16\textwidth}
    \begin{subfigure}{\textwidth}
    \includegraphics[width=\textwidth]{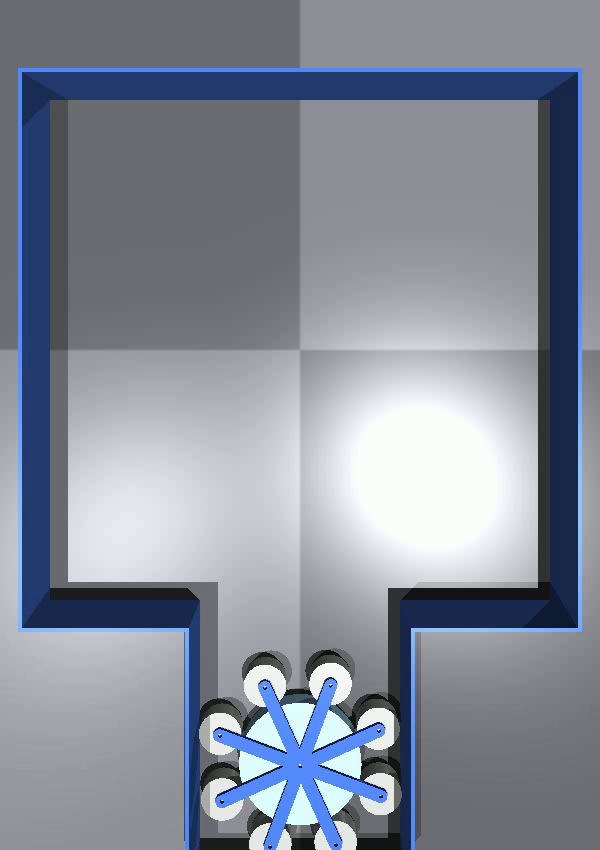}
    \caption{200s}
    \end{subfigure}
\end{minipage}
\caption{Robots move object without physical connection}
\label{fig: simulation: robot move object without physical connection}
\end{figure}

The goal is to move the object to the target location which is the bottom smaller space. The robots were controlled by Alg.~\ref{alg: move with connection}. As shown in Fig.~\ref{fig: simulation: robot move object without physical connection}, the object was moved to the target location. 

\subsubsection{Move object without connection}
\begin{figure}[h!]
\hfill
\begin{minipage}[b]{0.3\textwidth}
    \begin{subfigure}{\textwidth}
    \includegraphics[width=\textwidth]{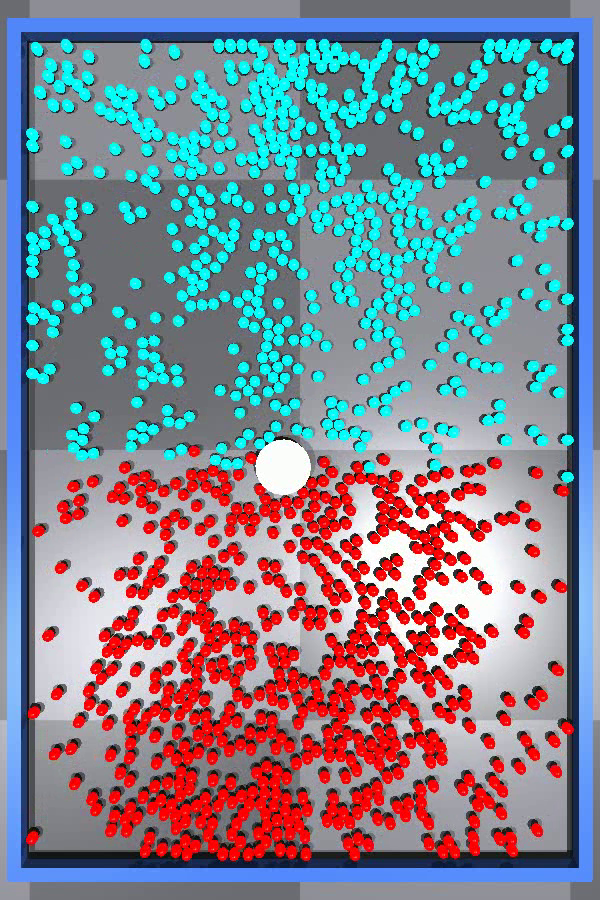}
    \caption{Simulation 3: move object}
    \label{fig: simulation: experimental settings of 1200 robot move object}
    \end{subfigure}
\end{minipage}
\hfill
\begin{minipage}[b]{0.3\textwidth}
    \begin{subfigure}{\textwidth}
    \includegraphics[width=\textwidth]{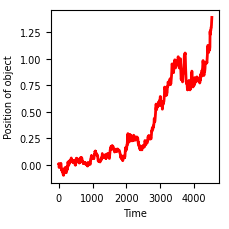}
    \caption{Exp3: intelligence experiments}
    \label{fig: simulation: intelligence experiments 1200, curve plot}
    \end{subfigure}
\end{minipage}
\caption{1200 robots moving object without physical connection}
\end{figure}

The goal is to move the object to the top of the constrained space. The robots in this experiment were controlled by Alg.~\ref{alg: move without connection}. As shown in Fig.~\ref{fig: simulation: experimental settings of 1200 robot move object}, the white object was placed in the center and the goal was to move the white object from the center to the top. The vertical position of the white object was plotted in Fig.~\ref{fig: simulation: intelligence experiments 1200, curve plot}.

\subsection{Robotic experiments}
\subsubsection{Robots design}
\begin{figure}
    \centering
    \includegraphics[width=\textwidth]{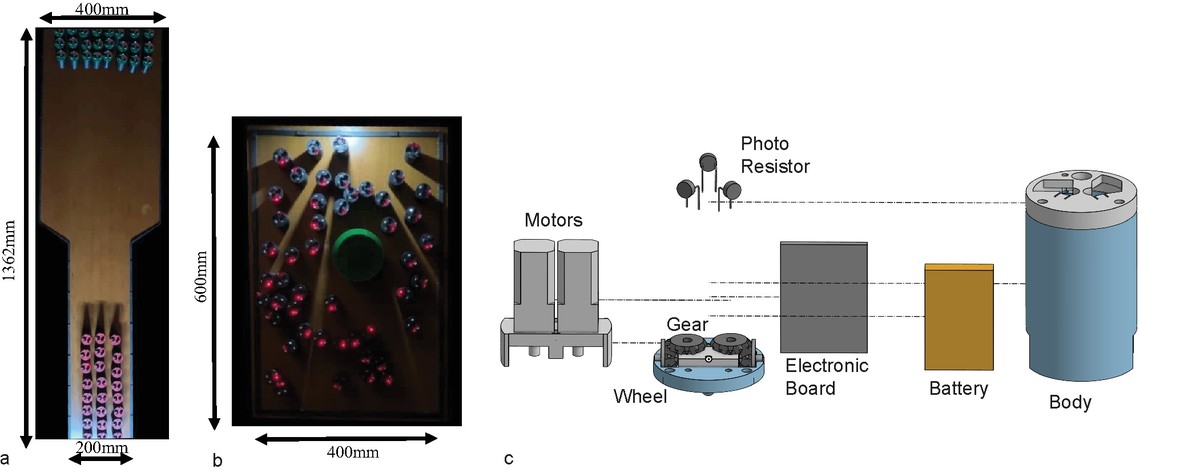}
    \caption{Robot experiments. (a) The dimension of moving object experiment. (b) The dimension of crossing experiment. (c) The exploration view of the robot hardware.}
    \label{fig: robot experimental settings}
\end{figure}

Each robot consists of two motors, control board, battery, gears, wheels, body, and photoresistors with an exploded view in Fig.~\ref{fig: robot experimental settings}c. The robot dimension is illustrated in Tab.~\ref{tab:robot hardware config} and its utilities are listed in Tab.~\ref{tab:robot utility configurations}. The positions of the photoresistors are illustrated in Fig.~\ref{fig: robot experimental settings}c.
\begin{table}[!h]
    \centering
    \begin{tabular}{c|c}
    \hline
    Name                        &  Dimension(mm)\\
    \hline
    Robot Height                & 60mm        \\
    Robot Diameter              & 33mm        \\
    Wheel diameter              & 12mm         \\
    Bevel gear diameter         & 12mm         \\
    Diameter of moving object   & 100mm        \\
    \hline
    \end{tabular}
    \caption{Robot hardware dimension and configurations}
    \label{tab:robot hardware config}
\end{table}
\begin{table}[!h]
    \centering
    \begin{tabular}{c|c|c}
    \hline
    Part name & configuration & number \\\hline
    Battery   & Lithium Polymer Battery 3.7V 250mAh 502030 &1   \\
    Motor     & DC,3V,N20, 52RPM &2    \\
    Bevel Gear & &2     \\
    photoresistor& GL5516 Photo Light Sensitive Resistor &3 \\
    \hline
    \end{tabular}
    \caption{Robots utility configurations}
    \label{tab:robot utility configurations}
\end{table}

\subsubsection{Robot experimental settings}
The configuration of the narrow road is shown in Fig.~\ref{fig: robot experimental settings}a. The long-duration test of robots crossing each other through a narrow road is illustrated in Fig.~\ref{fig: long duration of robot crossing}. When the robots reached the opposite end of the road, they were manually moved to the original side to test long-term operational robustness. The total duration is 1.55 hours when reaching the limit of battery storage. 
The dimension of experiments with moving objects is illustrated in Fig.~\ref{fig: robot experimental settings}b. 

\bibliography{references}
\bibliographystyle{Science}

\end{document}